%% file: main.tex
\documentclass[runningheads]{llncs}

\usepackage[mobile]{eccv}

\usepackage[utf8]{inputenc}
\usepackage[T1]{fontenc}

\usepackage[nolist]{acronym}
\usepackage{algorithm}
\usepackage{algpseudocode}
\usepackage{amsmath,amsfonts,amssymb}
\usepackage[accsupp]{axessibility}
\usepackage{booktabs}
\usepackage{cite}
\usepackage{eccvabbrv}
\usepackage{enumitem}
\usepackage{graphicx}
\usepackage{microtype}
\usepackage{pifont}
\usepackage{pgfplots}
\usepackage{relsize}
\usepackage[subrefformat=parens]{subcaption}
\usepackage{tikz}
\usepackage[normalem]{ulem}
\usepackage[dvipsnames]{xcolor}
\usepackage{xspace}

\usepackage[breaklinks,colorlinks,citecolor=eccvblue]{hyperref}
\usepackage[capitalize, nameinlink]{cleveref}
\usepackage{orcidlink}

\pgfplotsset{compat=1.18}

\input{acronyms}
\input{commands}

\begin{document}

\title{Hyperion -- A fast, versatile symbolic Gaussian Belief Propagation framework for Continuous-Time SLAM} 
\titlerunning{Hyperion -- A fast, versatile continuous-time GBP framework}
\author{David Hug\inst{1}\orcidlink{0000-0002-4430-3877} \and
Ignacio Alzugaray\inst{2}\orcidlink{0000-0002-7121-0000} \and
Margarita Chli\inst{1,3}\orcidlink{0000-0001-5611-7492}}
\authorrunning{D.~Hug et al.}

\institute{Vision for Robotics Lab, ETH Zürich, Zürich, Switzerland \and
 Dyson Robotics Lab, Imperial College London, London, United Kingdom \and
 Vision for Robotics Lab, University of Cyprus, Nicosia, Cyprus}

\maketitle
\begin{abstract}
  \input{sections/abstract}
  \keywords{Gaussian Belief Propagation \and Continuous-Time SLAM \and Distributed Non-Linear Least Squares Optimization \and B- and Z-Splines}
\end{abstract}
\section{Introduction}
\label{sec:introduction}
\input{sections/introduction}
\section{Related Work}
\label{sec:related_work}
\input{sections/related_work}
\section{Methodology}
\label{sec:methodology}
\input{sections/methodology}
\section{Experiments}
\label{sec:experiments}
\input{sections/experiments}
\section{Conclusions}
\label{sec:conclusions}
\input{sections/conclusions}
\section*{Acknowledgments}
\label{sec:acknowledgments}
\input{sections/acknowledgments}
\bibliographystyle{splncs04}
\bibliography{bibliography}

\end{document}

%% file: acronyms.tex
\begin{acronym}
    \acro{ADMM}{Alternating Direction Method of Multipliers}
    \acro{BA}{Bundle Adjustment}
    \acro{CPU}{Central Processing Unit}
    \acro{CTSLAM}{Continuous-Time \acl{SLAM}}
    \acro{DGR}{Dual Gaussian Representation}
    \acro{DTSLAM}{Discrete-Time \acl{SLAM}}
    \acro{DoF}{Degree of Freedom}
    \acro{DVS}{Dynamic Vision Sensor}
    \acro{GBP}{Gaussian Belief Propagation}
    \acro{GNSS}{Global Navigation Satellite System}
    \acro{GPS}{Global Positioning System}
    \acro{IMU}{Inertial Measurement Unit}
    \acro{INS}{Inertial Navigation System}
    \acro{KLT}{Kanade–Lucas–Tomasi}
    \acro{LIDAR}{Light Detection And Ranging}
    \acro{MAV}{Micro Aerial Vehicle}
    \acro{MGR}{Mixed Gaussian Representation}
    \acro{MH}{Machine Hall}
    \acro{MLE}{Maximum \textit{a posteriori} Likelihood Estimation}
    \acro{MoCap}{Motion Capture}
    \acro{NLLS}{Non-Linear Least Squares}
    \acro{RMSE}{Root Mean Square Error}
    \acro{RTK}{Real-Time Kinematic Positioning}
    \acro{SLAM}{Simultaneous Localization And Mapping}
    \acro{UWB}{Ultra Wide Band}
    \acro{VR}{Vicon Room}
\end{acronym}

%% file: commands.tex
\newcommand*{\hyperion}{\href{https://github.com/VIS4ROB-lab/hyperion}{Hyperion}\@\xspace}

\newcommand{\Romannumeral}[1]{\uppercase\expandafter{\romannumeral #1\relax}}

\newlength\longest

\DeclareSymbolFont{bbold}{U}{bbold}{m}{n}
\DeclareSymbolFontAlphabet{\mathbbold}{bbold}

\newcommand*{\via}{\textit{via}\@\xspace}

\algblock{Input}{EndInput}
\algnotext{EndInput}
\algblock{Output}{EndOutput}
\algnotext{EndOutput}

\newcommand{\cmark}{\ding{51}}
\newcommand{\xmark}{\ding{55}}
\DeclareMathOperator*{\argmax}{\arg\!\max}
\DeclareMathOperator*{\argmin}{\arg\!\min}

\newcommand*{\hBold}[1]{\ensuremath{\boldsymbol{#1}}}
\newcommand*{\hBracket}[1]{\ensuremath{\left(#1\right)}}
\newcommand*{\hNorm}[2][]{\ensuremath{{\left\lVert#2\right\rVert}_{#1}}}

\newcommand*{\hMat}[1]{\begin{bmatrix}#1\end{bmatrix}}

\newcommand*{\hR}[1][]{\ensuremath{\mathbb{R}^{#1}}}

\newcommand*{\hCov}{\hBold{\Sigma}}
\newcommand*{\hPre}{\hBold{\Lambda}}
\newcommand*{\hEst}[1]{\ensuremath{\hat{#1}}}

\newcommand*{\hRes}{\hBold{r}}

\newcommand*{\SU}[1]{\ensuremath{\mathbb{SU}(#1)}}
\newcommand*{\SO}[1]{\ensuremath{\mathbb{SO}(#1)}}
\newcommand*{\SE}[1]{\ensuremath{\mathbb{SE}(#1)}}

\newcommand*{\World}{w}
\newcommand*{\Body}{b}
\newcommand*{\Sensor}{s}
\newcommand*{\hSensors}{\mathcal{S}}

\newcommand*{\Basis}[1]{\ensuremath{\mathcal{B}_{#1}}}

\newcommand*{\Est}[1]{\ensuremath{\hat{#1}}}
\newcommand*{\Qx}[2]{\ensuremath{\boldsymbol{q}_{{#1}{#2}}}}
\newcommand*{\Rx}[2]{\ensuremath{\boldsymbol{R}_{{#1}{#2}}}}
\newcommand*{\Tx}[3][]{\ensuremath{{}^{#1}\boldsymbol{t}_{{#2}{#3}}}}
\newcommand*{\Tf}[2]{\ensuremath{\boldsymbol{T}_{{#1}{#2}}}}
\newcommand*{\Px}[1]{\ensuremath{\boldsymbol{p}_{{#1}}}}
\newcommand*{\EstQx}[2]{\ensuremath{\Est{\boldsymbol{q}}_{{#1}{#2}}}}
\newcommand*{\EstTx}[3][]{\ensuremath{{}^{#1}\Est{\boldsymbol{t}}_{{#2}{#3}}}}

\newcommand*{\EstTf}[2]{\ensuremath{\Est{\boldsymbol{T}}_{{#1}{#2}}}}
\newcommand*{\EstPx}[1]{\ensuremath{\Est{\boldsymbol{p}}_{{#1}}}}

\newcommand*{\Qwb}{\Qx{\World}{\Body}}

\newcommand*{\Qws}{\Qx{\World}{\Sensor}}

\newcommand*{\EstQws}{\EstQx{\World}{\Sensor}}

\newcommand*{\Rwb}{\Rx{\World}{\Body}}

\newcommand*{\twb}[1][]{\Tx[#1]{\World}{\Body}}

\newcommand*{\tws}[1][]{\Tx[#1]{\World}{\Sensor}}

\newcommand*{\Esttws}[1][]{\EstTx[#1]{\World}{\Sensor}}

\newcommand*{\Twb}{\Tf{\World}{\Body}}

\newcommand*{\Tbs}{\Tf{\Body}{\Sensor}}
\newcommand*{\Tsb}{\Tf{\Sensor}{\Body}}
\newcommand*{\Tws}{\Tf{\World}{\Sensor}}

\newcommand*{\EstTwb}{\EstTf{\World}{\Body}}
\newcommand*{\EstTbw}{\EstTf{\Body}{\World}}

\newcommand*{\EstTws}{\EstTf{\World}{\Sensor}}

\newcommand*{\hMeasurement}{\hBold{m}}
\newcommand*{\hMeasurements}{\mathcal{M}_s}
\newcommand*{\hMetric}{\hBold{\mu}}

%% file: sections/abstract.tex
\ac{CTSLAM} has become a promising approach for fusing asynchronous and multi-modal sensor suites. Unlike discrete-time \acs{SLAM}, which estimates poses discretely, \ac{CTSLAM} uses continuous-time motion parametrizations, facilitating the integration of a variety of sensors such as rolling-shutter cameras, event cameras and \acp{IMU}. However, \ac{CTSLAM} approaches remain computationally demanding and are conventionally posed as centralized \ac{NLLS} optimizations. Targeting these limitations, we not only present the fastest SymForce-based \cite{Martiros:etal:RSS22} B- and Z-Spline implementations achieving speedups between 2.43x and 110.31x over Sommer \etal \cite{Sommer:etal:CVPR2020} but also implement a novel continuous-time \ac{GBP} framework, coined \hyperion, which targets decentralized probabilistic inference across agents. We demonstrate the efficacy of our method in motion tracking and localization settings, complemented by empirical ablation studies.\\
\noindent \textbf{Code:}\;\;\url{https://github.com/VIS4ROB-lab/hyperion}

%% file: sections/introduction.tex
Estimating a sensor-suite's ego-motion and workspace employing \ac{SLAM} techniques, has long been studied using a wide variety of sensing modalities, such as vision sensors \cite{Wang:etal:ICCV2017, Schubert:etal:ECCV2018, Qin:etal:ARX2019b, Yang:etal:ICRA2021}, \acp{IMU} \cite{Qin:etal:TRO2017, Campos:etal:TRO2021, Karrer:ICRA2021}, \ac{GPS} feeds \cite{Lynen:etal:IROS2013, Mascaro:etal:ICRA2018}, and 
laser ranging \cite{Droeschel:ICRA2018} sensors. Unlike traditional, discrete-time approaches, which require careful synchronization of sensory measurements due to the discretization of motion-parametrizing states, \acf{CTSLAM} \cite{Oth:etal:CVPR2013, Hug:etal:RAL2022, Furgale:etal:ICRA2012, Furgale:etal:IROS2013, Droeschel:ICRA2018} offers native support for the fusion of asynchronous measurements due to its continuous-time parametrization that yields pose, velocity and acceleration estimates at arbitrary instances in time.

Despite their advantages, \ac{CTSLAM} approaches often entail higher computational complexity than conventional approaches, which somewhat hinders their deployment in real-world scenarios. In addition, most (discrete- and continuous-time) \ac{SLAM} systems pose the underlying optimization as a centralized \acf{NLLS} problem, which, without further modifications, strictly limits their applicability to single-agent setups.

\begin{figure}[t]
    \centering
    \subfloat[At initialization]{
      \includegraphics[height=3.4cm,trim={33cm 9cm 25cm 9cm},clip]{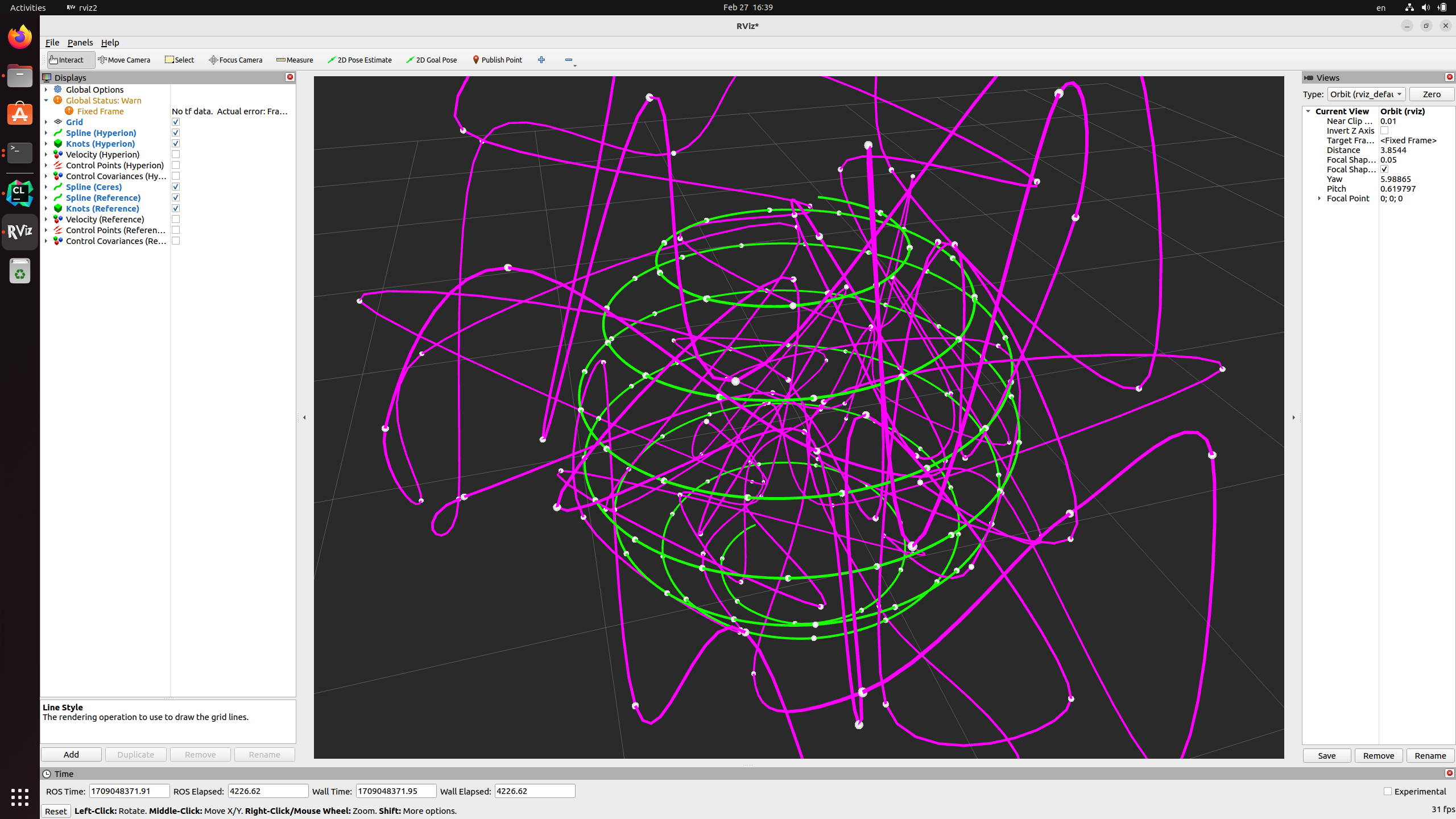}
    } 
    \subfloat[After the 1\textsuperscript{st} iteration]{
      \includegraphics[height=3.4cm,trim={33cm 9cm 25cm 9cm},clip]{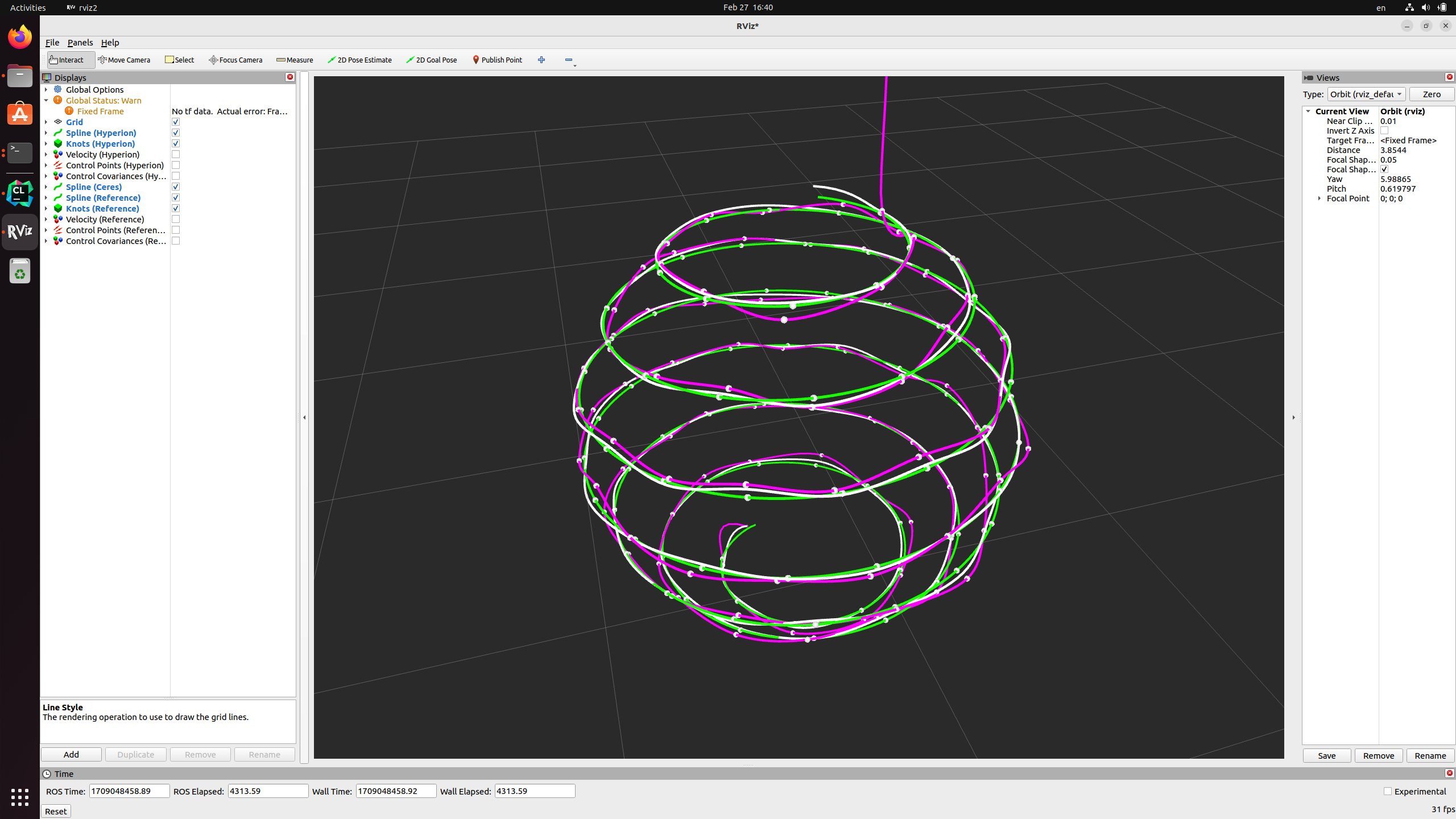}
    }
    \subfloat[Absolute Error]{
      \resizebox{!}{3.4cm}{\input{tables/absolute_rmse}}
    }
    \caption{Both the proposed continuous-time \ac{GBP} solver (in magenta) and the conventional \ac{NLLS} solver \cite{Agarwal:etal:Ceres} (in white) converge to identical solutions close to the ground truth (in green) even under poor initialization ($\pm 1.00$ m/rad) and substantial pose measurement noise ($\pm 0.05$ m/rad).}
    \label{fig:eccv2024:teaser}
\end{figure}

Thus, novel, decentralized algorithms along with more effective continuous-time motion parametrizations are paramount in the quest to overcome existing challenges in traditional, centralized \ac{CTSLAM} techniques. Distributed methods such as \ac{GBP} are especially promising given that they, in contrast to conventional \ac{NLLS} paradigms, achieve iterative, probabilistic inference through message-passing between individual nodes and factors in a factor graph and that they operate in a distributed and asynchronous manner by nature rendering them inherently scalable, even across multiple agents. In addition, \ac{GBP} also explicitly models the uncertainties of optimizable quantities, which can be leveraged to selectively direct computational resources to non-converged nodes in the graph. This circumstance is especially promising in the context of \ac{CTSLAM} where effective allocation of computational resources towards the least certain estimates promises to further reduce their computational complexity.

This work, in particular, pushes the performance envelope of B- and Z-Splines and devises an innovative, distributed optimization strategy for \ac{CTSLAM} using \ac{GBP}. Whilst the need for faster continuous-time parametrizations is self-evident, we also specifically target the absence of a distributed, continuous-time state estimation framework. Such a framework, not only promises flexible resource allocation and decentralized state estimation but also allows fusing asynchronous measurements. Here, we showcase the practicality of the proposed continuous-time \ac{GBP} framework and provide it as an open-source implementation to encourage benchmarking. In addition to these contributions, this work also
\renewcommand{\labelitemi}{$\bullet$}
\begin{itemize}
    \item offers the fastest, fully analytic B- and Z-Spline implementations to date,
    \item presents a novel, symbolic \ac{GBP}-based framework for continuous-time \acs{SLAM},
    \item demonstrates the suitability of the proposed method in practical setups, and
    \item provides detailed ablation studies on the algorithm itself.
\end{itemize}

%% file: tables/absolute_rmse.tex
\begin{tikzpicture}
    \pgfplotsset{
        scale only axis,
        xmin=0, xmax=9.9
    }
    \begin{axis}[
      axis y line*=left,
      xlabel=Time,
      ylabel={Absolute Rotation error [rad]},
      legend image post style={black},
      legend style = {text=black,font=\footnotesize},
    ]
    \legend{Ours, Ceres}
    \addplot[very thick,smooth,dotted] table [x=t, y=dr, col sep=comma] {tables/hyperion_absolute_rmse.csv};
    \addplot[very thick,smooth] table [x=t, y=dr, col sep=comma] {tables/ceres_absolute_rmse.csv};
    \end{axis}
    \begin{axis}[
      blue,
      axis y line*=right,
      axis x line=none,
      ylabel={Absolute Translation error [m]}
    ]
    \addplot[very thick,smooth,dotted] table [x=t, y=dt, col sep=comma] {tables/hyperion_absolute_rmse.csv};
    \addplot[very thick,smooth] table [x=t, y=dt, col sep=comma] {tables/ceres_absolute_rmse.csv};
    \end{axis}
\end{tikzpicture}

%% file: sections/related_work.tex
The \ac{SLAM} problem has long been researched as it comprises the core of robotic perception, with seminal works focusing on discrete-time, monocular \cite{MurArtal:etal:TRO2015, Qin:etal:TRO2017} and stereo setups \cite{Qin:etal:ARX2019, Qin:etal:ARX2019b}, as well as works which propose fusion of additional sensing information, such as inertial \cite{Leutenegger:etal:IJRR2015, Bloesch:etal:IROS2015, Bloesch:etal:IJRR2017, Campos:etal:TRO2021}, and laser ranging \cite{Droeschel:ICRA2018}. In addition to these feature-based approaches to \ac{SLAM}, alternative powerful paradigms have been proposed, such as direct methods \cite{Engel:etal:ECCV2014, Forster:etal:TRO2017} and machine-learning based techniques \cite{Milford:ICRA2012, Wang:etal:ICRA2017} have been demonstrated to be beneficial in some scenarios.

In contrast to traditional, discrete-time \ac{SLAM} methods that have been widely adopted, \ac{CTSLAM} techniques have early on been demonstrated to exhibit great potential for high-fidelity and continuous estimates of motion \cite{Lovegrove:etal:BMVC2013, Furgale:etal:ICRA2012, Furgale:etal:IROS2013, Anderson:ICRA2014, Mueggler:etal:TRO2018, Wang:etal:ACCV2018, Usenko:etal:RAL2019, Hug:etal:RAL2022}, albeit posing fundamental scientific and algorithmic challenges. The key advantage of these methods lies in their inherent capability to fuse unsynchronized and asynchronous measurements (\eg from a rolling shutter camera or an event-based vision sensor) in the estimation processes, in contrast to conventional approaches. However, the wider adoption of \ac{CTSLAM} techniques remains impeded by challenges in finding more efficient motion representations, addressing convergence issues as well as overcoming computational limitations, missing out on the promise for high-fidelity motion and scene estimation.

Following promising leads in fundamental \ac{SLAM} research for single agents, a novel challenge and desire to deploy the same techniques in collaborative, multi-agent setups arose. Works such as \cite{Schmuck:JFR2019} proposed means to address centralized multi-agent \ac{SLAM}, while \cite{Karrer:ICRA2021} investigated the advantages of variable-stereo baseline setups, which leverage the views from two agents to boost the accuracy of high-altitude depth estimates. A common approach to solve \ac{NLLS} optimizations in a distributed fashion is based on \ac{ADMM} approaches similar to the work in \cite{Peng:etal:SISC2016} and \cite{Banninger:etal:ICRA2023}, which leverage the use of dual residuals to ensure consistent estimates across distributed multi-agent \ac{NLLS} optimizations. Orthogonally, \ac{GBP} approaches \cite{Davison:ARX2019, Ortiz:etal:CVPR2020, Ortiz:etal:ARX2021, Murai:etal:TRO2023} can perform distributed and asynchronous inference of states via an equivalent message-passing scheme, making them ideal candidates to address multi-agent \ac{SLAM}.

In this work, we build upon these ideas, consolidate a distributed \ac{GBP} optimization with a continuous-time parametrization, and demonstrate the suitability of the novel, combined method in practical setups.

%% file: sections/methodology.tex
\subsection{Preliminaries}
\label{sec:eccv2024:preliminaries}
All \acf{SLAM} algorithms, in essence, aim to estimate an optimal, cost-minimizing set of optimizable parameters $\hBold{\Theta}$ which are set to describe a collection of noisy sensory measurements $\hMeasurement$ with the highest possible accuracy. Specifically, in its \ac{NLLS} formulation, this optimal solution is found by minimization of the cumulative cost over a set of weighted residuals $\hBold{\bar{r}}$ associated with measurements $\hMeasurement$ stemming from a connected sensor $s$. A weighted residuals $\hBold{\bar{r}}$ at measurement time $t$ is computed by comparing a predicted measurement $\hEst{\hMeasurement}(t,\hBold{\theta}_s)$ with sensor-associated parameters $\hBold{\theta}_s\subseteq\hBold{\Theta}$ to a measured one $\hMeasurement(t)$ through the application of a metric $\hMetric$ (\ie $\hEst{x} \boxminus_{\hMetric} x$) and a subsequent weighting \via the square-root information $\hBold{\Omega}_m$ according to
\begin{gather}
    \label{eq:eccv2024:residual}
    \hRes(t, \hBold{\theta}_s) = \hEst{\hMeasurement}(t, \hBold{\theta}_s) \boxminus_{\hMetric} \hMeasurement(t)~\text{and}\\
    \label{eq:eccv2024:weighted_residual}
    \hNorm{\hBold{\bar{r}}}^2 = \hBold{\bar{r}}^\top \hBold{\bar{r}} = \hRes^\top \hBold{\Omega}_m^\top \hBold{\Omega}_m \hRes = \hRes^\top \hPre_m \hRes = \hRes^\top \hCov_m^{-1} \hRes,
\end{gather}
where $\hCov_m$ and $\hPre_m$ are the covariance and precision matrix, respectively. Ultimately, we aim to minimize the sum over residuals stemming from all connected sensors $\hSensors$ and their associated measurements $\hMeasurements$ with measurement times $\mathcal{T}_s$. To this end, established \ac{NLLS} solvers \cite{Agarwal:etal:Ceres} are commonly deployed in practice, which aim to obtain the optimal, cost-minimizing parameters $\hBold{\Theta}^\ast$ as defined below.
\begin{equation}
    \label{eq:eccv2024:nlls}
    \hBold{\Theta}^\ast = \underset{\hBold{\Theta}}{\argmin}\left[\mathlarger{\sum}_{s\in\mathcal{S}} \mathlarger{\sum}_{t\in\mathcal{T}_s}~\frac{1}{2}~\hNorm{\hBold{\bar{r}}(t, \hBold{\theta}_s)}^2\right].
\end{equation}

\subsection{Continuous-Time Motion}
\label{sec:eccv2024:continuous_time_motion}
\begin{figure}[t]
    \centering
    \includegraphics[width=0.6\linewidth]{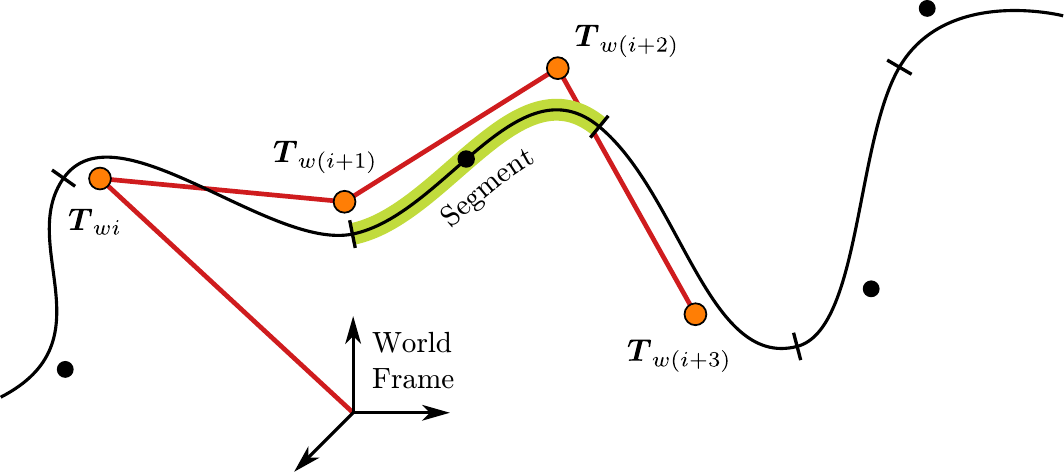}
    \caption{For every (valid) instance in time $t$, a collection of adjacent bases (in orange) gives rise to an individual segment (in green) of a cubic B-Spline. An interpolated pose at query time $t$ is then obtained from the (cumulative) blending of these bases.}
    \label{fig:eccv2024:continuous_parametrization}
\end{figure}
In recent times, continuous-time formulations of conventional \ac{SLAM} algorithms have been studied by multiple authors \cite{Furgale:etal:ICRA2012, Furgale:etal:IROS2013, Lovegrove:etal:BMVC2013, Hug:3DV2020, Sommer:etal:CVPR2020, Hug:etal:RAL2022}, where the most common motion parametrization is based on the concept of cubic B-Splines. This practice is rooted in several favorable properties of B-Splines, such as their compact representation, finite support, analytic Jacobians, and their $\mathcal{C}_2$-continuity for predicting instantaneous velocities and accelerations. In this work, we utilize split interpolations for world-to-body transformations $\Twb(t)\in\SE3$, illustrated in \cref{fig:eccv2024:continuous_parametrization}, that separately parametrize rotations $\Rwb(t)\in\SO3$ and translations $\twb(t)\in\hR[3]$. In particular, we compute transformations $\Twb(t)$ using
\begin{equation}
    \label{eq:eccv2024:T_wb}
    \Twb\hBracket{t} = \hMat{\Rwb\hBracket{\Qwb\hBracket{t}} & \twb\hBracket{t} \\ \hBold{0} & 1} \in\SE3~\text{with}
\end{equation}
\begin{equation}
    \label{eq:eccv2024:q_wb}
    \Qwb\hBracket{t} = \Qx{w}{i} * \prod_{j=1}^{k} \hBracket{\Qx{w}{\hBracket{i+j-1}}^{-1} * \Qx{w}{\hBracket{i+j}}}^{\lambda_{j}(t)}
\end{equation}
\begin{equation}
    \label{eq:eccv2024:t_wb}
    \twb\hBracket{t} = \Tx{w}{i} + \sum_{j=1}^{k} \left[\lambda_{j}(t)\hBracket{\Tx{w}{\hBracket{i+j}} - \Tx{w}{\hBracket{i+j-1}}}\right],
\end{equation}
where individual evaluations of $\Twb\left(t\right)$ depend on a collection of bases $\{\Basis{i},\dots,\Basis{i+k}\}$ (see \cref{fig:eccv2024:continuous_parametrization}) and each basis $\Basis{i}$ comprises a time $t_i$, a quaternion $\Qx{w}{i}$ and a translation $\Tx{w}{i}$. Above, $k$ is the \ac{DoF} of the B-/Z-Spline, rotations $\Rwb(t)$ are parametrized by quaternions $\Qwb(t)$ and the expression $\lambda_{j}(t)$ serves as a placeholder for concrete interpolation coefficients found in \cite{Qin:PCCGA1998, Becerra:SAM2003, Lovegrove:etal:BMVC2013, Hug:etal:RAL2022}.

\subsection{Continuous-Time Optimization}
\label{sec:eccv2024:continuous_time_optimization}
In its probabilistic formulation \cite{Ortiz:etal:CVPR2020, Ortiz:etal:ARX2021}, the non-linear minimization problem from \cref{eq:eccv2024:nlls} is equivalent to finding an optimal probability distribution that accounts for all acquired measurements. Without loss of generality, one can also represent the problem as a product of factors $f_i\propto e^{-E_i(\hBold{\theta}_i)}$ with induced energies $E_i(\hBold{\theta}_i)$, arriving at the modified expression to obtain the optimal parameters $\hBold{\Theta}^\ast$.
\begin{gather}
    \label{eq:eccv2024:p_nlls_1}
    \hBold{\Theta}^\ast = \underset{\hBold{\Theta}}{\argmax} \log\left(p\left(\hBold{\Theta}\right)\right) = \underset{\hBold{\Theta}}{\argmin} \sum_i E_i(t_i, \hBold{\theta}_i)~\text{with}\\
    \label{eq:eccv2024:p_nlls_2}
    p\left(\hBold{\Theta}\right) = \prod_i f_i\left(t_i, \hBold{\theta}_i\right) \propto \prod_i e^{-E_i\left(t_i, \hBold{\theta}_i\right)}.
\end{gather}
In this work, we use multi-variant Gaussians $\mathcal{N}(\hBold{\mu}_i, \hCov_i)$ to model the factors $f_i$ in \cref{eq:eccv2024:p_nlls_2}, linking the energies in \cref{eq:eccv2024:p_nlls_1} to the residuals in \cref{eq:eccv2024:weighted_residual} according to
\begin{gather}
    \label{eq:eccv2024:energy_function}
    E_i(t_i, \hBold{\theta}_i) = \hNorm{\hBold{\bar{r}}_i}^2 = \hRes_i^\top(t_i, \hBold{\theta}_i) \hBold{\Omega}_i^\top \hBold{\Omega}_i \hRes_i(t_i, \hBold{\theta}_i).
\end{gather}
The generic residuals $\hBold{\bar{r}}_i$ themselves are non-linear with a corresponding Taylor expansions in $\hBold{\theta}_i$ around some linearization point $\hBold{\theta}_i^0$ such that
\begin{gather}
    \label{eq:eccv2024:linearized_residual}
    \hBold{\bar{r}}_i(\hBold{\theta}_i) - \hBold{\bar{r}}_i(\hBold{\theta}_i^0) \approx
    D \hBold{\bar{r}}_i(\hBold{\theta}_i^0)\,(\hBold{\theta}_i - \hBold{\theta}_i^0) =
    \hBold{\bar{J}}_i^0(\hBold{\theta}_i - \hBold{\theta}_i^0) = \hBold{\bar{J}}_i^0 \hBold{\tau}_i^0.
\end{gather}
Furthermore, all factors $f_i$, their energies $E_i(t_i, \hBold{\theta}_i)$ respectively, can also be converted to an equivalent, incremental information form $\mathcal{N}^{-1}(\hBold{\eta}_i^0, \hPre_i^0)$, yielding
\begin{gather}
    \label{eq:eccv2024:E_i_inf}
    E_i(t_i, \hBold{\tau}_i^0) \approx \frac{1}{2}\:\hBold{\tau}_i^{0,\top} \hPre_i^0 \hBold{\tau}_i^0 - \hBold{\tau}_i^{0,\top} \hBold{\eta}_i^0 \quad\text{where}\\
    \label{eq:eccv2024:eta_and_lambda_i_inf}
    \hBold{\eta}_i^0 = -\hBold{\bar{J}}_i^{0,\top} \hBold{\bar{r}}_i^0
    \quad\text{and}\quad
    \hPre_i^0 = \hBold{\bar{J}}_i^{0,\top} \hBold{\bar{J}}_i^0.
\end{gather}
This dual representation of factors $f_i$ and the existence of the Gaussian $\mathcal{N}$ and its inverse $\mathcal{N}^{-1}$ is vital for efficient conditioning and marginalization of individual terms in the \ac{GBP} algorithm presented in the next section.
\begin{figure}[t]
    \centering
    \subfloat[Factor Graph]{
        \includegraphics[height=2.75cm]{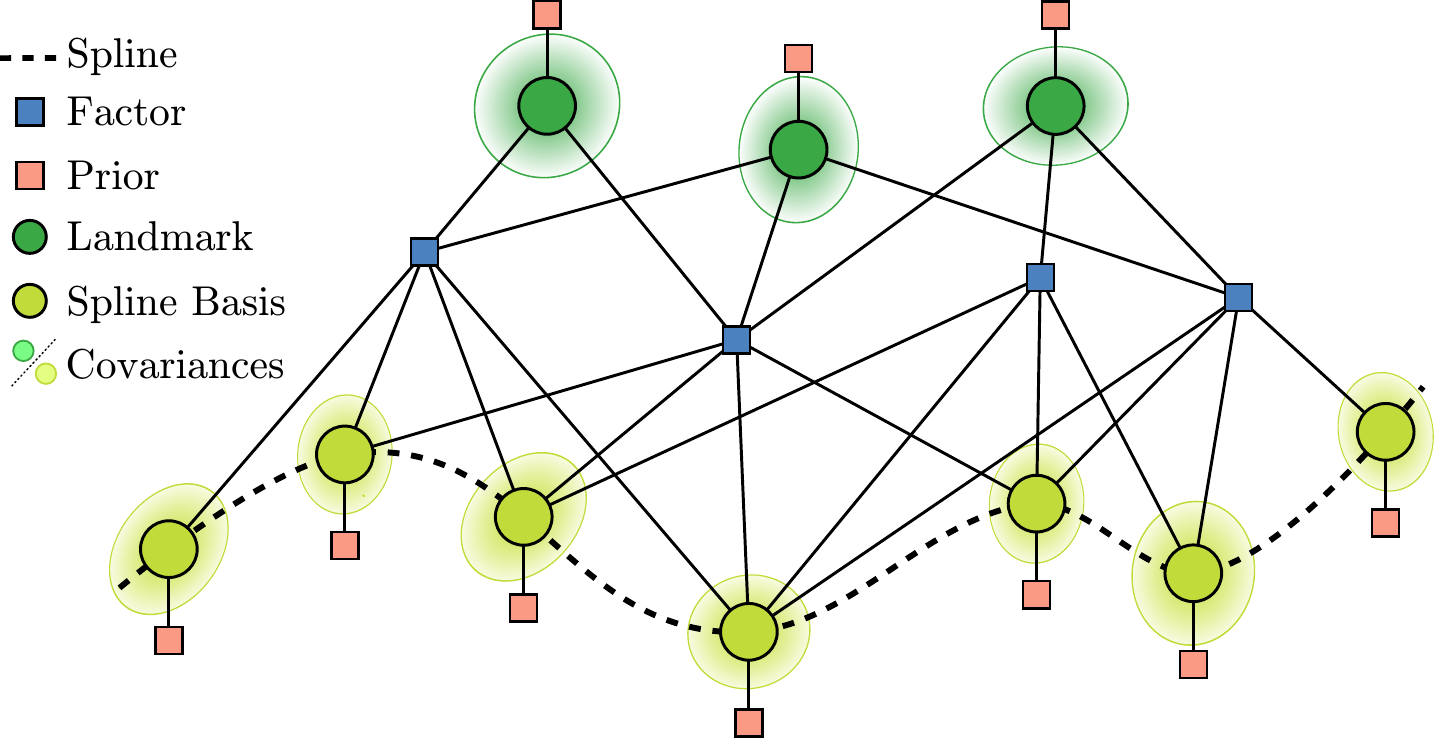}
        \label{fig:eccv2024:factor_graphs}
    } 
    \subfloat[Message Passing]{
        \includegraphics[height=2.75cm]{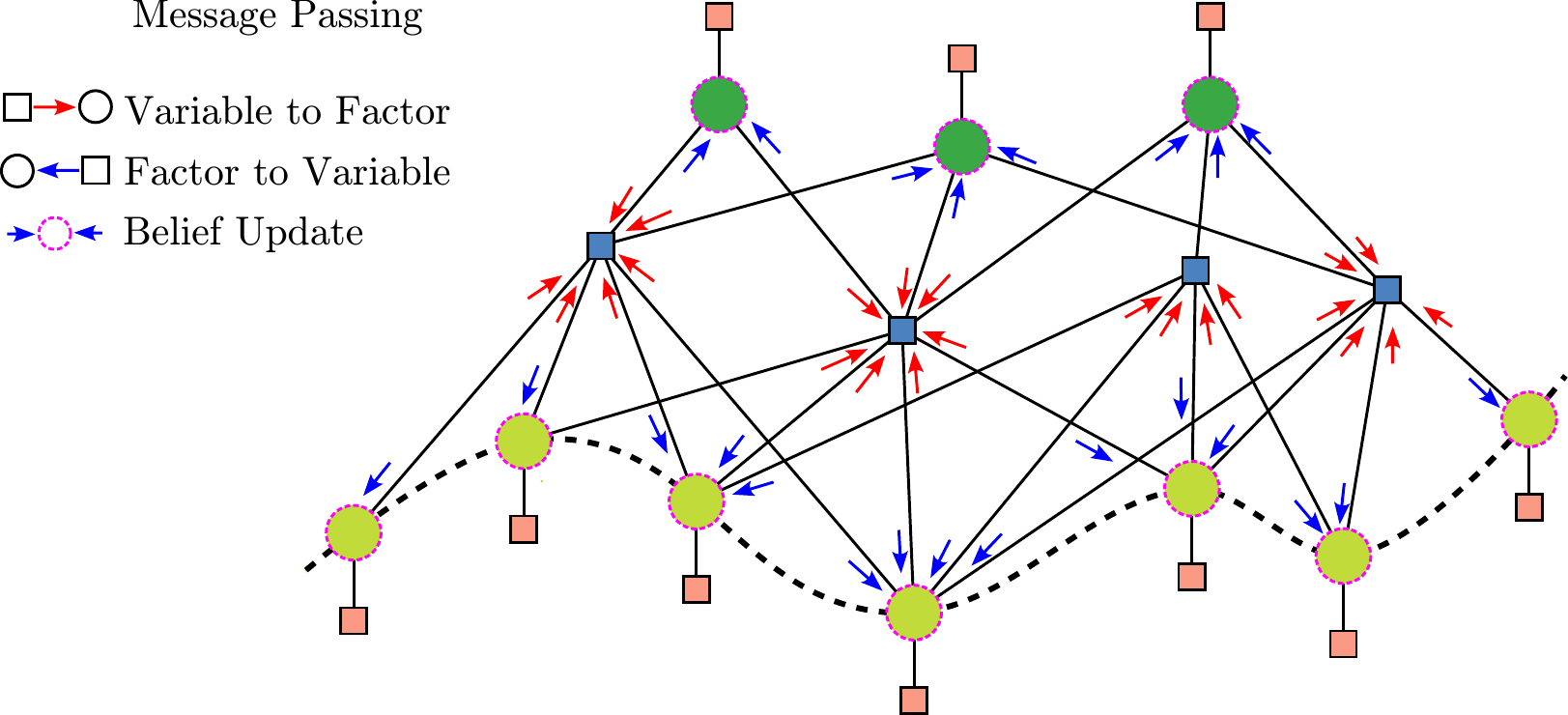}
        \label{fig:eccv2024:message_passing}
    }
    \caption{
        \subref{fig:eccv2024:factor_graphs} Qualitative illustration of the factor graphs resulting from a continuous-time motion parametrization. Notably, even one of the simplest factor archetypes, which purely relies on the motion-parameterizing nodes as well as some landmarks, introduces a considerable amount of loops in the continuous-time realm. This can be attributed to the fact that poses $\Twb(t)$ depend on multiple bases for any given time $t$.
        \subref{fig:eccv2024:message_passing} Visualization of the message passing algorithm between nodes and factors in a graph $\mathcal{G}$.
    }
\end{figure}

\subsubsection{\acf{GBP}}
\label{sec:eccv2024:gaussian_belief_propagation}
As illustrated in \cref{fig:eccv2024:factor_graphs}, \cref{eq:eccv2024:nlls,eq:eccv2024:p_nlls_1} expose equivalent visual representations as factor graphs, giving rise to compact and intuitive descriptions of complex probabilistic dependencies which are encoded into bipartite graphs $\mathcal{G}$ containing factors $f_i\sim\mathcal{N}^{-1}(\hBold{\eta}_{f_i}, \hPre_{f_i})$ and nodes $n_j\sim \mathcal{N}(\hBold{\mu}_{n_j}, \hCov_{n_j}) = \mathcal{N}^{-1}(\hBold{\eta}_{n_j}, \hPre_{n_j})$. Intuitively, the \ac{GBP} algorithm, thus, comprises two steps to iteratively solve for the optimal solution to \cref{eq:eccv2024:p_nlls_1}, namely node and factor updates with intermediate factor-to-node $\hBold{m}_{f_i\rightarrow n_j}$ and node-to-factor $\hBold{m}_{n_j\rightarrow f_i}$  message passing detailed in the following.

\paragraph{Node Updates:}
\label{sec:eccv2024:node_updates}
The Gaussian nodes $n_j\in\mathcal{G}$ with neighborhood $N(n_j)$ of connected factors $f_i \in N(n_j)$ are updated by taking the product over incoming factor-to-node messages $\hBold{m}_{f_i \rightarrow n_j}$, which reduces to a simple summation of Gaussians for (linear) vector spaces, resulting in node beliefs $B(n_j) = \mathcal{N}^{-1}(\hBold{\eta}_{n_j}, \hPre_{n_j})$, where
\begin{gather}
    \label{eq:eccv2024:linear_node_update}
    \hBold{\eta}_{n_j} = \hBold{\eta}_{n_j}^p + \sum_{f_i\in N(n_j)} \hBold{\eta}_{f_i \rightarrow n_j}
    \quad\text{and}\quad
    \hPre_{n_j} = \hPre_{n_j}^p + \sum_{f_i\in N(n_j)} \hBold{\Lambda}_{f_i \rightarrow n_j}.
\end{gather}
Above, $P(n_j) = \mathcal{N}^{-1}(\hBold{\eta}_{n_j}^p, \hPre_{n_j}^p)$ denotes a known prior on the node belief $B(n_j)$ which can be leveraged to better constrain the problem in practice. However, despite its simplicity and elegance, this straightforward approach can not be applied to Lie groups $G$ with associated tangent space $\mathfrak{g}$ and their operators $\boxminus:G\times G\mapsto\mathfrak{g}$ and $\boxplus:G\times\mathfrak{g}\mapsto G$ extensively used in robotics. Hence, a more sophisticated approach is required to obtain the product of factor-to-node messages and to handle frame conversions between elements in the Lie group.
%
%
%

To this end, we follow a similar strategy as Murai \etal\cite{Murai:etal:TRO2023} and propose the use of a \ac{MGR} to parameterize Lie-group-valued nodes as $n_j \sim \mathfrak{N}(\hBold{\mu}_{n_j}, \hPre_{n_j})$, where $\hBold{\mu}_{n_j} \in G$ and $\hPre_{n_j} \in \hR[\dim(\mathfrak{g}) \times \dim(\mathfrak{g})]$, also assuming that incoming message $\hBold{m}_{f_i\rightarrow n_j} \sim \mathfrak{N}(\hBold{\mu}_{f_i\rightarrow n_j}, \hPre_{f_i\rightarrow n_j})$ take the same form. In contrast to conventional vector spaces, the precision matrices are expressed relative to their associated elements in the Lie group, implying that one must warp them to a privileged frame of reference. An intuitive choice for such a frame is the latest estimate of the node's state, denoted as $\hBold{\mu}_{n_j}^0$ and $\hPre_{n_j}^0$, resulting in the following transformation rules to warp the messages.
\begin{gather}
    \label{eq:eccv2024:lie_node_update_0}
    \hBold{\tau}_{f_i\rightarrow n_j}^0 = \hBold{\mu}_{f_i\rightarrow n_j} \boxminus \hBold{\mu}_{n_j}^0 \quad\in\hR[\dim(\mathfrak{g})] \\
    \label{eq:eccv2024:lie_node_update_1}
    \hPre_{f_i\rightarrow n_j}^0 =
    \left[\frac{\partial\hBold{\tau}_{f_i\rightarrow n_j}^0}{\partial\hBold{\mu}_{f_i\rightarrow n_j}}\right]^\top \hPre_{f_i\rightarrow n_j} \left[\frac{\partial\hBold{\tau}_{f_i\rightarrow n_j}^0}{\partial\hBold{\mu}_{f_i\rightarrow n_j}}\right] \quad\in\hR[\dim(\mathfrak{g})\times\dim(\mathfrak{g})]
\end{gather}
After this conversion, both $\hBold{\tau}_{f_i\rightarrow n_j}^0$ and $\hPre_{f_i\rightarrow n_j}^0$ are elements in the tangent (matrix) space relative to the privileged frame and must then be summed in analogy to \cref{eq:eccv2024:linear_node_update} to yield the intermediate, incremental values
\begin{gather}
    \label{eq:eccv2024:lie_node_update_2}
    \hBold{\tau}_{n_j}^+ = \alpha_{n_j}\sum_{f_i\in N(n_j)} \hPre_{f_i\rightarrow n_j}^0 \hBold{\tau}_{f_i\rightarrow n_j}^0 \quad\text{and}\quad
    \hPre_{n_j}^+ = \sum_{f_i\in N(n_j)} \hPre_{f_i\rightarrow n_j}^0,
\end{gather}
where $\alpha_{n_j}$ denotes an optional step size. The above increments must then again be warped to the updated frame of reference by evaluating
\begin{gather}
    \label{eq:eccv2024:lie_node_update_3}
    \hBold{\mu}_{n_j} = \hBold{\mu}_{n_j}^0 \boxplus \hBold{\tau}_{n_j}^+
    \quad\text{and}\quad
    \hPre_{n_j} =
    \left[\frac{\partial\hBold{\mu}_{n_j}}{\partial\hBold{\tau}_{n_j}^+}\right]^\top
    \hPre_{n_j}^+
    \left[\frac{\partial\hBold{\mu}_{n_j}}{\partial\hBold{\tau}_{n_j}^+}\right].
\end{gather}
Note, however, that priors can not be injected as simplistically as in \cref{eq:eccv2024:linear_node_update} and need to be treated as proper factors in the context of Lie groups instead.

\paragraph{Node-to-factor Messages:}
\label{sec:eccv2024:node_to_factor_message}
The generation of node-to-factor messages mirrors the expressions from \cref{eq:eccv2024:lie_node_update_2}, differing only in excluding information stemming from the target factor in the sums. Hence, they encompass
\begin{gather}
    \label{eq:eccv2024:node_to_factor_messages_0}
    \hBold{\tau}_{n_j \rightarrow f_k}^+ = \sum_{f_i\in N(n_j) \setminus f_k} \hPre_{f_i\rightarrow n_j}^0 \hBold{\tau}_{f_i\rightarrow n_j}^0 \quad\text{and}\quad
    \hPre_{n_j \rightarrow f_k}^+ = \sum_{f_i\in N(n_j) \setminus f_k} \hPre_{f_i\rightarrow n_j}^0
\end{gather}
alongside their respective linearization point $\hBold{\mu}_{n_j}^0$ to form the outgoing message triplet $(\hBold{\mu}_{n_j}^0, \hBold{\tau}_{n_j \rightarrow f_k}^+, \hPre_{n_j \rightarrow f_k}^+)$ destined for the factor $f_k$. Note, however, that another reevaluation of \cref{eq:eccv2024:lie_node_update_0,eq:eccv2024:lie_node_update_1} is required to generate outgoing messages that leverage the latest state estimate from \cref{eq:eccv2024:lie_node_update_3}.

\paragraph{Factor Updates:}
\label{sec:eccv2024:factor_updates}
In analogy to previous paragraphs, factors $f_i\in\mathcal{G}$ depend on a collection of connected, neighboring nodes $N(f_i)$ which collectively determine the linearization point $\hBold{\theta}_{f_i}^0$ for the residual evaluation $\hBold{\bar{r}}_{f_i}$ from \cref{eq:eccv2024:weighted_residual}. The computation of the factor-to-node messages $\hBold{m}_{f_i\rightarrow N(f_i)}$ depends on the factor beliefs $B(f_i) = \mathcal{N}^{-1}(\hBold{\eta}_{f_i}^0, \hPre_{f_i}^0)$ obtained from \cref{eq:eccv2024:eta_and_lambda_i_inf} as well as the auxiliary, intermediate quantities $\hBold{\eta}_{f_i}^\prime$ and $\hBold{\Lambda}_{f_i}^\prime$ (used in the next paragraph) defined as
\begin{gather}
    \label{eq:eccv2024:eta_lambda_fi_prime}
    \hBold{\eta}_{f_i}^\prime  = \hBold{\eta}_{f_i}^0 + \hBold{\eta}_{N(f_i)\rightarrow f_i}^+
    \quad\text{and}\quad
    \hBold{\Lambda}_{f_i}^\prime = \hBold{\Lambda}_{f_i}^0 + \hBold{\Lambda}_{N(f_i)\rightarrow f_i}^+
\end{gather}
Above, $\hBold{\eta}_{N(f_i)\rightarrow f_i}^+$ and $\hBold{\Lambda}_{N(f_i)\rightarrow f_i}^+$ denote stacked vector and block diagonal matrix versions of the neighborhood-to-factor messages (see \cref{eq:eccv2024:eta_fi_prime,eq:eccv2024:lambda_fi_prime}).

\paragraph{Factor-to-Node Messages:}
\label{sec:eccv2024:factor_to_node_message}
We illustrate the computation of the factor-to-node messages (see \cref{fig:eccv2024:message_passing}) assuming a factor $f_i$ which depends on two nodes, namely $n_a$ and $n_b$. Thus, \cref{eq:eccv2024:eta_lambda_fi_prime} takes the following form
\begin{gather}
    \label{eq:eccv2024:eta_fi_prime}
    \hBold{\eta}_{f_i}^\prime =
    \hMat{
        \hBold{\eta}_{a}^\prime \\
        \hBold{\eta}_{b}^\prime \\
    } =
    \hBold{\eta}_{f_i}^0 + \hMat{
        \hBold{\eta}_{n_a\rightarrow f_i}^+ \\
        \hBold{\eta}_{n_b\rightarrow f_i}^+ \\
    } \\
    \label{eq:eccv2024:lambda_fi_prime}
    \hBold{\Lambda}_{f_i}^\prime =
    \hMat{
        \hBold{\Lambda}_{aa}^\prime & \hBold{\Lambda}_{ba}^{\prime \top} \\
        \hBold{\Lambda}_{ba}^\prime & \hBold{\Lambda}_{bb}^\prime
    } =
    \hBold{\Lambda}_{f_i}^0 + \hMat{
        \hBold{\Lambda}_{n_a\rightarrow f_i}^+ & \hBold{0} \\
        \hBold{0} & \hBold{\Lambda}_{n_b\rightarrow f_i}^+ \\
    }.
\end{gather}
To retrieve the message $\hBold{m}_{f_i\rightarrow n_a} = \mathfrak{N}(\hBold{\mu}_{f_i \rightarrow n_a}, \hPre_{f_i \rightarrow n_a})$, the remaining nodes (\ie $n_b$ in this example) must initially be marginalized by computing the Schur complement defined as 
\begin{gather}
    \label{eq:eccv2024:eta_a_message}
    \hBold{\eta}_{f_i \rightarrow n_a}^\prime = \hBold{\eta}_{a}^0 - \hBold{\Lambda}_{ba}^{\prime \top} \hBold{\Lambda}_{bb}^{\prime -1} \hBold{\eta}_{b}^\prime
    \quad\text{and}\quad
    \hBold{\Lambda}_{f_i \rightarrow n_a}^\prime = \hBold{\Lambda}_{aa}^0 - \hBold{\Lambda}_{ba}^{\prime \top} \hBold{\Lambda}_{bb}^{\prime -1} \hBold{\Lambda}_{ba}^\prime.
\end{gather}
In a similar vein, one obtains $\hBold{m}_{f_i\rightarrow n_b}$ by permuting $\hBold{\eta}_{f_i}^\prime$ and $\hBold{\Lambda}_{f_i}^\prime$ such that $n_a$ is marginalized instead of $n_b$. In particular,
\begin{gather}
    \label{eq:eccv2024:schur_complement}
    \hMat{
        \hBold{\eta}_{b}^\prime \\
        \hBold{\eta}_{a}^\prime \\
    } = P \hBold{\eta}_{f_i}^\prime
    ~\text{and}~
    \hMat{
        \hBold{\Lambda}_{bb}^\prime & \hBold{\Lambda}_{ab}^{\prime \top} \\
        \hBold{\Lambda}_{ab}^\prime & \hBold{\Lambda}_{aa}^\prime
    } = P \hBold{\Lambda}_{f_i}^\prime P^\top,
\end{gather}
which also naturally extends to factors touching more than two nodes. Readers might notice that our formulation differs from others found in the literature; this has the advantage that the formulations from \cref{eq:eccv2024:eta_fi_prime,eq:eccv2024:lambda_fi_prime} allow for extremely efficient in-place permutations of $\hBold{\eta}_{f_i}^\prime$ and $\hBold{\Lambda}_{f_i}^\prime$ to marginalize many-node factors, limiting reallocations and recomputations. Mirroring the process in \cref{eq:eccv2024:lie_node_update_3}, it is necessary to convert the incremental values $\hBold{\eta}_{f_i \rightarrow n_a}^\prime$ and $\hBold{\Lambda}_{f_i \rightarrow n_a}^\prime$ into the updated, outgoing frame of reference, to obtain the message $\hBold{m}_{f_i\rightarrow n_a}$, relying on an optional step size $\alpha_{f_i}$. That is
\begin{gather}
    \label{eq:eccv2024:factor_to_node_message}
    \hBold{\tau}_{f_i\rightarrow n_a}^\prime = \alpha_{f_i} \hBold{\Lambda}_{f_i \rightarrow n_a}^{\prime -1}  \hBold{\eta}_{f_i \rightarrow n_a}^\prime
    ~\in\hR[\dim(\mathfrak{g})], \quad
    \hBold{\mu}_{f_i\rightarrow n_a} = \hBold{\mu}_{n_a}^0 \boxplus \hBold{\tau}_{f_i\rightarrow n_a}^\prime
    ~\in G \\
    \text{and}\quad
    \hBold{\Lambda}_{f_i\rightarrow n_a} =
    \left[\frac{\partial \hBold{\mu}_{f_i\rightarrow n_a}}{\partial \hBold{\tau}_{f_i\rightarrow n_a}^\prime}\right]^\top \hBold{\Lambda}_{f_i\rightarrow n_a}^\prime
    \left[\frac{\partial \hBold{\mu}_{f_i\rightarrow n_a}}{\partial \hBold{\tau}_{f_i\rightarrow n_a}^\prime}\right]
    ~\in\hR[\dim(\mathfrak{g})\times\dim(\mathfrak{g})].
\end{gather}

%

\paragraph{Robust Residuals and Energies:}
\label{sec:eccv2024:robust_residuals_and_energies}
It is well understood that the presence of outliers in the optimization problem from \cref{eq:eccv2024:nlls} causes substantial issues in terms of converging to a globally optimal solution $\hBold{\Theta^\ast}$ due to the quadratic nature of the occurring cost terms. The same holds in the context of \ac{GBP}, where outliers are bound to heavily influence the found solution as well. To address this issue we take inspiration from established \ac{NLLS} approaches \cite{Agarwal:etal:Ceres} and apply robust loss functions $\rho$ to \cref{eq:eccv2024:nlls,eq:eccv2024:energy_function}, resulting in modified expressions for the robust energies $\breve{E}$ and the robust optimal solution $\hBold{\breve{\Theta}^\ast}$, provided for completeness only.
\begin{equation}
    \label{eq:eccv2024:robust_energy}
    \breve{E}(t, \hBold{\theta}_s) = \frac{1}{2}~\rho\left(\hBold{\bar{r}}^\top \hBold{\bar{r}}\right)
\end{equation}
\begin{equation}
    \hBold{\breve{\Theta}}^\ast = \underset{\hBold{\Theta}}{\argmin}\left[\mathlarger{\sum}_{s\in\mathcal{S}} \mathlarger{\sum}_{t\in\mathcal{T}_s}~\breve{E}(t, \hBold{\theta}_s)\right]
\end{equation}
In particular, based on the methodology from Triggs \etal\cite{Triggs:etal:VA2000}, the robust residual and Jacobian then take the following forms.
\begin{gather}
    \alpha^2 - 2 \alpha - \frac{2 \rho''}{\rho'} \hBold{\bar{r}}_i^\top \hBold{\bar{r}}_i \overset{!}{=} 0 \\
    \hBold{\breve{r}}_i =
    \frac{\sqrt{\rho'}}{1-\alpha} \hBold{\bar{r}}_i
    \quad
    \hBold{\breve{J}}_i =
    \sqrt{\rho'} \left(1 - \alpha\:\frac{\hBold{\bar{r}}_i \hBold{\bar{r}}_i^\top}{\hNorm{\hBold{\bar{r}}_i}^2}\right)\hBold{\bar{J}}_i
\end{gather}

\paragraph{Non-Gaussian Nodes:}
\label{sec:eccv2024:non_gaussian_nodes}
%
Another extension to conventional \ac{GBP} lies in classifying Gaussian nodes as either pure constants or as stochastic variables. This distinction streamlines computation given that the message marginalization $\hBold{m}_{n_j\rightarrow f_i}$ across nodes \via Cholesky decomposition exposes complexity $\mathcal{O}(n^3)$ which quickly becomes prohibitive. In particular, allowing non-variable parameters (\eg extrinsic, intrinsic, offset \etc) to be treated as pure constants considerably reduces the dimensionality of the marginalization procedure without any drawbacks.

\subsection{Sensor Models}
\label{sec:eccv2024:sensor_models}
In the following, we detail the employed absolute and visual sensor models, where expressions marked with $\hat{\cdot}$ indicate optimizable parameters. Conversely, unless explicitly mentioned, all other variables are considered known and constant.
\subsubsection{Absolute Sensor Model}
\label{sec:eccv2024:absolute_sensor_model}
The predictions of absolute sensor measurements $\hEst{\hMeasurement}_s(t_m,\Theta_s)$ (\eg absolute pose estimates extracted from AprilTags \cite{Olson:ICRA2011} or measurements obtained from a \ac{MoCap} system) are evaluated using \cref{eq:eccv2024:absolute_measurement} and subsequently compared against their corresponding true measurements by applying the metric $\boxminus_{\hMetric}$ in \cref{eq:eccv2024:absolute_metric} where we make use of the logarithmic map for elements in $\SO3$, and $\SU2$ respectively.
\begin{gather}
    \label{eq:eccv2024:absolute_measurement}
    \hat{\hMeasurement}_s(t_m, \Theta_{s}) = \EstTws(t_m) = \EstTwb(t_m)\:\Tbs \\
    \label{eq:eccv2024:absolute_metric}
    \EstTws\boxminus_{\hMetric} \Tws = \hMat{
        \log\left(\EstQws\:\Qws^{-1}\right) \\
        \Esttws - \tws
    }\in\hR[6]
\end{gather}
\subsubsection{Visual Sensor Model}
\label{sec:eccv2024:visual_sensor_model}
Abstracting from the specific camera parameters, such as its intrinsics and its distortion model, we introduce the mapping $\pi(\cdot)$ in \cref{eq:eccv2024:visual_measurement} to infer individual projections of landmarks $l_w$ onto the image plane at time $t_m$. For visual, pixel-based measurements, the metric $\boxminus_{\hMetric}$ is chosen to be equivalent to trivial Euclidean subtraction. 
\begin{gather}
    \label{eq:eccv2024:visual_measurement}
    \hEst{\hMeasurement}_s(t_m, \Theta_{s}) = \EstPx{s}(t_m) = \pi(\Tsb\EstTbw(t_m),\: l_w) \\
    \EstPx{s} \boxminus_{\hMetric} \Px{s} = \EstPx{s} - \Px{s} \in\hR[2]
\end{gather}

\subsection{System}
\label{sec:eccv2024:system}
\subsubsection{Overview}
\label{sec:eccv2024:system_overview}
Our system is designed to efficiently support both batch and windowed optimizations, crucial for applications such as \ac{BA} and \ac{SLAM}. Inspired by Ceres \cite{Agarwal:etal:Ceres}, it allows dynamic modification of nodes and factors between solver calls, features multiple solving strategies, namely synchronous and dropout, determining the order of node and factor updates and leverages multi-threading to exploit the inherent parallelism of \ac{GBP} methods, further boosting the overall performance. Optimizable parameters are added and removed in analogy to Ceres \cite{Agarwal:etal:Ceres}, with the notable distinction that \ac{GBP}-based methods necessitate both an initial mean and covariance estimate for each parameter, unlike standard \ac{NLLS} methods. This requirement, however, poses no practical hindrance as conservative guesses on the covariance suffice to bootstrap the algorithm.
\subsubsection{Symbolic Factors}
\label{sec:eccv2024:system_symbolic_factors}
Targeting the long-standing challenge of obtaining efficient, analytical expressions and derivatives of spline-based motion parameterizations and their associated cost factors, here, we leverage SymForce \cite{Martiros:etal:RSS22}, a symbolic code generation framework, to completely automate this cumbersome, time-intensive and error-prone process. Taking inspiration from Sommer \etal \cite{Sommer:etal:CVPR2020}, who exploited Lie-group-specific properties to simplify spline-related mathematical expressions, we supercharge their recursive spline formulation by combining it with SymForce \cite{Martiros:etal:RSS22} to obtain ultra-efficient C++ cost factor implementations. Furthermore, recognizing the broader relevance of automated factor generation for (continuous-time) robotics applications, our framework offers a comprehensive library featuring spline-based and standard factors designed to seamlessly interoperate with both our framework and Ceres \cite{Agarwal:etal:Ceres}, facilitating widespread adoption.
\subsubsection{Update Strategies}
\label{sec:eccv2024:update_strategies}
Amongst many possible update strategies, we focus our attention on synchronous and dropout updates. Synchronous updates closely follow traditional \ac{NLLS} algorithms, where residuals are completely re-evaluated in every solver iteration and where (non-constant) nodes and factors are sequentially updated in an alternating fashion to ensure optimal convergence. In contrast to traditional approaches, \ac{GBP}, however, also accommodates selective updates based on a dropout strategy. That is, one assigns probabilities, $d_n$ for nodes and $d_f$ for factors, dictating their likelihood of being updated in each iteration. Although this may slow convergence, it intentionally compensates for stability challenges in loopy graphs while mimicking real-world scenarios with imperfect communication channels and delayed messages.

%% file: sections/experiments.tex
We evaluate the proposed system, coined \hyperion, empirically through simulations in \ac{MoCap} (absolute) and localization settings, using a temporal interval of 0.1 seconds between adjacent spline bases along with the empirical step sizes $\alpha_{n_j} = \alpha_{f_i} = 0.7$ across all cases. The simulated trajectories span 10 seconds and mimic real-world conditions with appropriate initial perturbations, sensor acquisition rates, and measurement noise levels. Our analysis focuses on benchmarking \hyperion against Ceres \cite{Agarwal:etal:Ceres}, an established, centralized \ac{NLLS} solver, and providing in-depth ablation studies on different aspects of the framework itself.

\subsection{Absolute Setup}
We commence our analysis with the absolute sensor configuration, capturing measurements akin to the ones stemming from a \ac{MoCap} system, where individual factors model direct observations of the underlying motion without auxiliary parameters/nodes. As such, they embody the most fundamental category of constraints, suggesting that derived factors, such as landmark projections or relative measurements, merely extend these elements, rendering them ideal candidates to analyze key properties of the proposed approach. In the following, we assume an acquisition rate of 40 Hz matching the real-world specifications.

Our initial examination centers on comparing the convergence behavior of \hyperion against Ceres \cite{Agarwal:etal:Ceres} across varying levels of initial perturbations and measurement noise. To this end, both solvers are identically initialized to a modified ground truth motion, encompassing perturbations in rotations and translations. Furthermore, we make use of the synchronized vertex updates if not stated otherwise, aiming for a fair comparison against Ceres.

The qualitative outcomes of this setup are illustrated in \cref{fig:eccv2024:teaser}, demonstrating that both \hyperion and Ceres converge to identical solutions even under substantial initial perturbation and considerable measurement noise. These observations are quantified in \cref{tab:eccv2024:perturbation_tolerance,tab:eccv2024:noise_tolerance}, indicating that motion estimates obtained from \hyperion and Ceres closely align across all levels of perturbation and noise.

Message dropouts, mimicking imperfect communication, have also been shown to enhance convergence \cite{Ortiz:etal:ICRA2022,Murai:etal:RAL2024} in loopy graphs. Hence, we study their effect on the proposed \ac{GBP} framework, summarizing key insights in \cref{fig:eccv2024:absolute_dropout}, revealing that \hyperion consistently converges toward identical solutions across various dropout ratios. Despite this, they also influence the required iteration count (\ie, updates of all vertices), with every 10\% increase in dropouts entailing an additional 2-4 iterations until convergence is reached.

Furthermore, we find that different motion parameterizations, specifically cubic Z- and B-Splines \cite{Becerra:SAM2003,Qin:PCCGA1998}, also influence the convergence of the proposed approach, which we summarize in \cref{fig:eccv2024:absolute_convergence}. A notable difference between the traditional and our proposed solver lies in how motion parameterizations affect the convergence speed of the latter. Our findings suggest that Z-Splines, which are interpolating rather than approximating splines, typically yield better-conditioned solutions since their bases must lie on the motion estimate itself, implicitly constraining the space of possible solutions. Naturally, the centralized \ac{NLLS} solver sets a baseline for the required iteration count to resolve an optimization problem, serving as a comparative standard for our method. Thus, while \ac{GBP}-based approaches are inherently distributed, \cref{fig:eccv2024:absolute_convergence} indicates that they impose between 2 to 4 additional iterations to achieve convergence.

\begin{figure}[t]
    \centering
    \subfloat[Estimate]{
      \includegraphics[height=3.1cm,trim={31cm 8cm 23cm 8cm},clip]{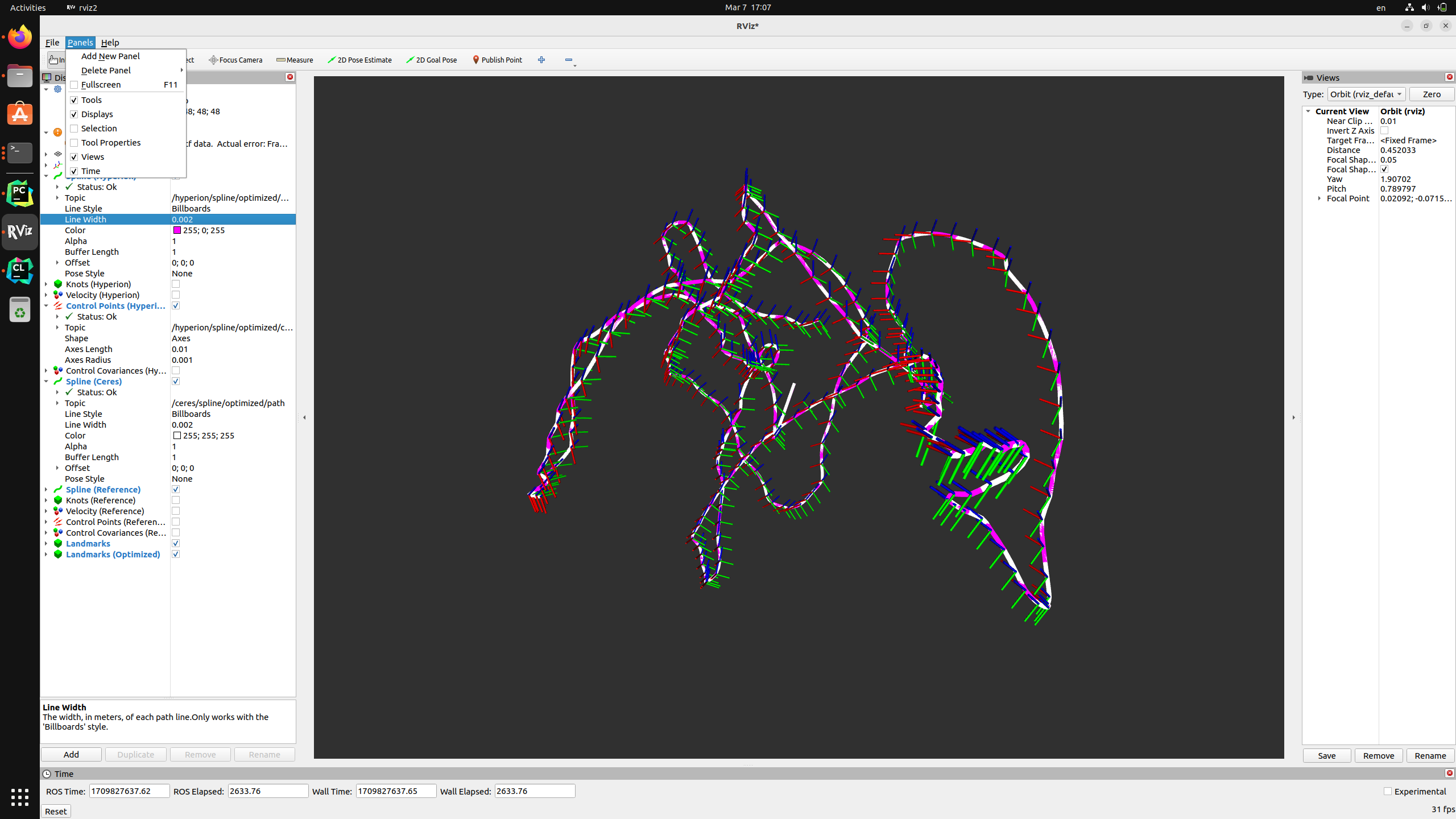}
      \label{fig:eccv2024:charuco_estimate}
    } 
    \subfloat[Convergence]{
      \resizebox{!}{3.1cm}{\input{tables/charuco_energy}}
      \label{fig:eccv2024:charuco_energy}
    }
    \subfloat[Relative Error]{
      \resizebox{!}{3.1cm}{\input{tables/charuco_error}}
      \label{fig:eccv2024:charuco_error}
    }
    \caption{ChArUco \cite{Garrido:etal:PR2014} setup with overlapping motion estimates \subref{fig:eccv2024:charuco_estimate} from \hyperion and Ceres \cite{Agarwal:etal:Ceres} in magenta and white, respectively. Illustration of the corresponding convergence \subref{fig:eccv2024:charuco_energy} and the relative errors between the two converged estimates \subref{fig:eccv2024:charuco_error}.}
    \label{fig:eccv2024:charuco_setup}
\end{figure}

Lastly, we benchmark \hyperion's real-world performance using a handheld camera to track a ChArUco \cite{Garrido:etal:PR2014} board, capturing a 60-second trajectory at 30 Hz with an iPhone 13 Pro. The analysis, detailed in \cref{fig:eccv2024:charuco_setup}, closely mirrors our previous analysis. However, we find that real-world motions, which innately expose a greater level of volatility than simulated ones, expose slower convergence.

\begin{table}[t]
\centering
\begin{minipage}{.45\textwidth}
\smaller
  \centering
  \begin{subtable}{\linewidth}
    \centering
    \resizebox{\columnwidth}{!}{
        \begin{tabular}{|c|cccccc|}
            \cline{2-7}
            \multicolumn{1}{c|}{} & \multicolumn{6}{c|}{\textbf{Perturbation [m/rad]}} \\
            \cline{2-7}
            \multicolumn{1}{c|}{} & 1e-5 & 1e-4 & 1e-3 & 1e-2 & 1e-1 & 1e-0 \\ 
            \cline{2-7} \noalign{\vspace{0.75ex}} \hline
            \multicolumn{7}{|c|}{Ours} \\
            \hline
            R [rad] & 5.2e-6 & 5.2e-6 & 5.2e-6 & 5.2e-6 & 5.9e-6 & 5.3e-6 \\ 
            t [m] & 5.8e-6 & 5.8e-6 & 5.9e-6 & 5.9e-6 & 6.0e-6 & 1.3e-5 \\
            \hline
            \multicolumn{7}{|c|}{Ceres} \\
            \hline
            R [rad] & 5.2e-6 & 5.2e-6 & 5.2e-6 & 5.2e-6 & 5.2e-6 & 5.2e-6 \\ 
            t [m] & 5.9e-6 & 5.9e-6 & 5.9e-6 & 5.9e-6 & 5.9e-6 & 5.9e-6 \\
            \hline
        \end{tabular}
    }
    \caption{Perturbation Tolerance}
    \label{tab:eccv2024:perturbation_tolerance}
  \end{subtable}
\end{minipage}%
\begin{minipage}{.45\textwidth}
\smaller
  \centering
  \begin{subtable}{\linewidth}
    \centering
    \resizebox{\columnwidth}{!}{
        \begin{tabular}{|c|cccccc|}
            \cline{2-7}
            \multicolumn{1}{c|}{} & \multicolumn{6}{c|}{\textbf{Noise [m/rad]}} \\
            \cline{2-7}
            \multicolumn{1}{c|}{} & 1e-5 & 1e-4 & 1e-3 & 1e-2 & 1e-1 & 1e-0 \\
            \cline{2-7} \noalign{\vspace{0.75ex}} \hline
            \multicolumn{7}{|c|}{Ours} \\
            \hline
            R [rad] & 5.2e-6 & 5.2e-5 & 5.2e-4 & 5.2e-3 & 5.3e-2 & 5.6e-1 \\ 
            t [m] & 5.8e-6 & 5.9e-5 & 5.9e-4 & 5.9e-3 & 5.9e-2 & 5.9e-1 \\
            \hline
            \multicolumn{7}{|c|}{Ceres} \\ 
            \hline
            R [rad] & 5.2e-6 & 5.2e-5 & 5.2e-4 & 5.2e-3 & 5.2e-2 & 5.5e-1 \\ 
            t [m] & 5.9e-6 & 5.9e-5 & 5.9e-4 & 5.9e-3 & 5.9e-2 & 5.9e-1 \\
            \hline
        \end{tabular}
    }
    \caption{Noise Tolerance}
    \label{tab:eccv2024:noise_tolerance}
  \end{subtable}
\end{minipage}
\caption{\acp{RMSE} in rotation (R) and translation (t) resulting from Ceres \cite{Agarwal:etal:Ceres} and \hyperion under different perturbation levels \subref{tab:eccv2024:perturbation_tolerance}, different noise levels \subref{tab:eccv2024:noise_tolerance} respectively. The solvers run to convergence or terminate after 50 iterations. The perturbation/noise setup assumes the lowest noise/perturbation, respectively.}
\end{table}
\begin{figure}[t]
    \centering
    \subfloat[Convergence: Dropout]{
        \resizebox{!}{3.6cm}{\input{tables/absolute_dropout}}
        \label{fig:eccv2024:absolute_dropout}
    } 
    \subfloat[B-Spline Evaluation Timings]{
    \begin{minipage}[c][4cm][c]{0.52\textwidth}
        \centering
        \resizebox{\columnwidth}{!}{
        \begin{tabular}{||ccc|cc|cc|cc|c||}
        \hline
        \multicolumn{3}{||c|}{Setup} & \multicolumn{2}{c|}{Pose [s]} & \multicolumn{2}{c|}{Velocity [s]} & \multicolumn{2}{c|}{Acceleration [s]} & Avg. \\
        $\mathcal{L}$ & $k$ & $\partial/\partial \mathcal{B}$ & Ours & Basalt & Ours & Basalt & Ours & Basalt & Speedup \\
        \hline
        \SO3 & 4 & \xmark & \textbf{1.64e-7} & 3.16e-7 & \textbf{9.49e-8} & 2.90e-7 & \textbf{1.12e-7} & 3.28e-7 & 2.64x \\
        \SO3 & 4 & \cmark & \textbf{5.03e-7} & 6.70e-6 & \textbf{4.05e-7} & 7.82e-6 & \textbf{5.11e-7} & 9.46e-6 & 17.05x \\
        \SO3 & 5 & \xmark & \textbf{1.96e-7} & 4.14e-7 & \textbf{1.32e-7} & 3.67e-7 & \textbf{1.39e-7} & 3.96e-7 & 2.58x \\
        \SO3 & 5 & \cmark & \textbf{6.89e-7} & 1.08e-5 & \textbf{5.82e-7} & 1.27e-5 & \textbf{7.87e-7} & 1.58e-5 & 19.19x \\
        \SO3 & 6 & \xmark & \textbf{2.17e-7} & 4.78e-7 & \textbf{1.82e-7} & 4.42e-7 & \textbf{1.81e-7} & 4.83e-7 & 2.43x \\
        \SO3 & 6 & \cmark & \textbf{8.19e-7} & 1.57e-5 & \textbf{7.51e-7} & 1.87e-5 & \textbf{1.01e-6} & 2.43e-5 & 22.71x \\
        \SE3 & 4 & \xmark & \textbf{1.62e-7} & 7.03e-7 & \textbf{1.38e-7} & 7.46e-7 & \textbf{1.34e-7} & 6.88e-7 & 4.96x \\
        \SE3 & 4 & \cmark & \textbf{5.71e-7} & 4.69e-5 & \textbf{9.11e-7} & 5.35e-5 & \textbf{1.12e-6} & 6.25e-5 & 65.56x \\
        \SE3 & 5 & \xmark & \textbf{1.96e-7} & 7.38e-7 & \textbf{1.91e-7} & 8.73e-7 & \textbf{1.70e-7} & 9.15e-7 & 4.57x \\
        \SE3 & 5 & \cmark & \textbf{7.32e-7} & 9.64e-5 & \textbf{1.27e-6} & 9.44e-5 & \textbf{1.47e-6} & 1.14e-4 & 94.53x \\
        \SE3 & 6 & \xmark & \textbf{2.53e-7} & 9.11e-7 & \textbf{2.34e-7} & 1.12e-6 & \textbf{2.23e-7} & 1.12e-6 & 4.47x \\
        \SE3 & 6 & \cmark & \textbf{9.29e-7} & 1.25e-4 & \textbf{1.54e-6} & 1.67e-4 & \textbf{1.99e-6} & 1.75e-4 & 110.31x \\
        \hline
        \end{tabular}
        }
        \end{minipage}
        \label{tab:eccv2024:timings}
    }
    \caption{
        \subref{fig:eccv2024:absolute_dropout} Graph energy vs. number of iterations conditioned on the dropout probability in the absolute setup (batch).
        \subref{tab:eccv2024:timings} Performance comparison between our symbolically, auto-generated and optimized B-Spline implementation and the recursively-defined, hand-crafted implementation used by Sommer \etal\cite{Sommer:etal:CVPR2020} on an M3 Max @4.05GHz.
    }
\end{figure}

\subsection{Localization Setup}
In this section, we analyze our approach in a localization setup, where we model an image acquisition rate of 20 Hz and 50 randomized, observable landmarks at a range of 2 to 6 m across the modeled motion, mimicking real-world conditions.

\begin{figure}[t]
    \centering
    \subfloat[At initialization]{
      \includegraphics[height=3.1cm,trim={35cm 12cm 23cm 8cm},clip]{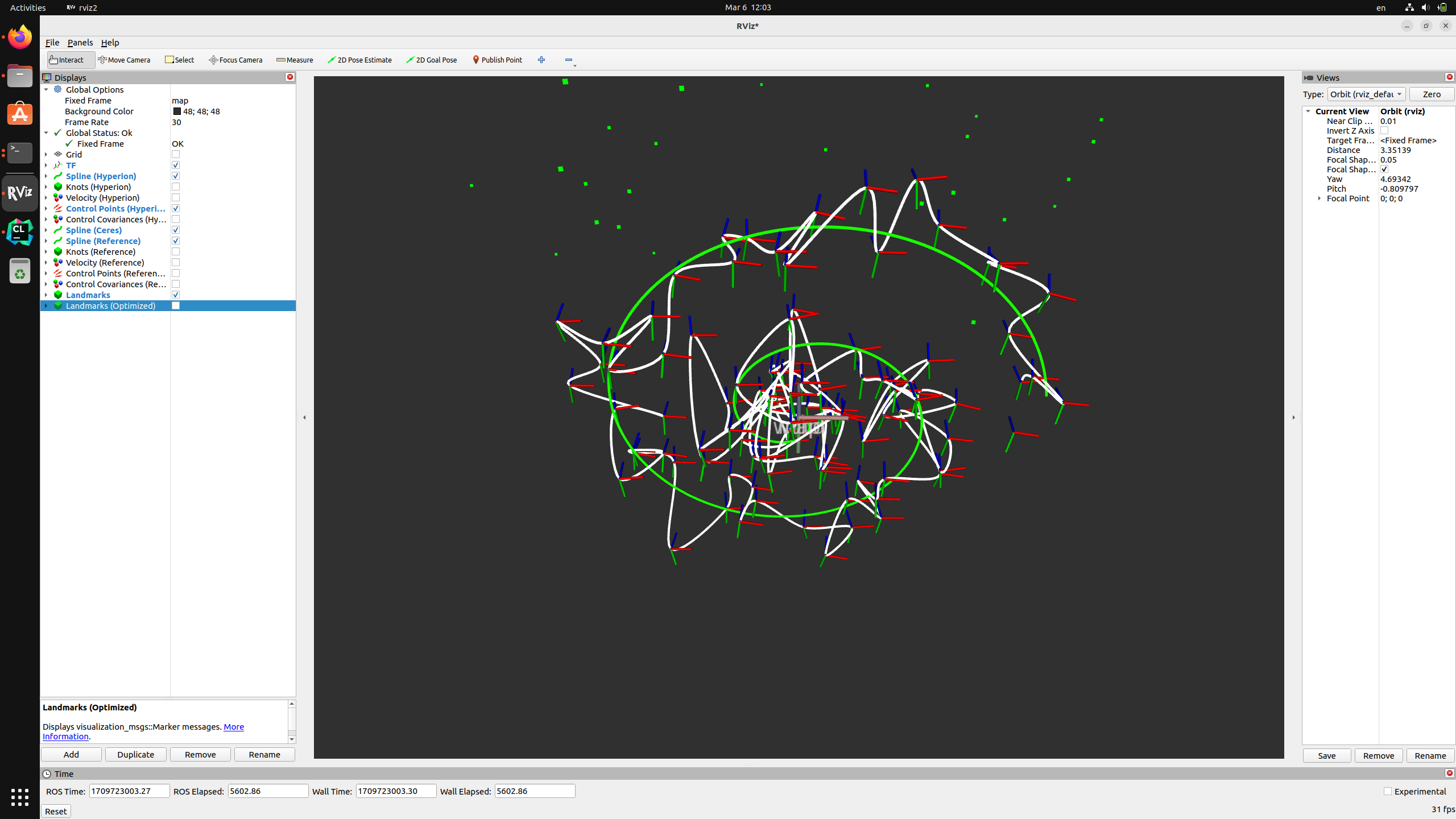}
    } 
    \subfloat[After the 1\textsuperscript{st} iteration]{
      \includegraphics[height=3.1cm,trim={35cm 12cm 23cm 8cm},clip]{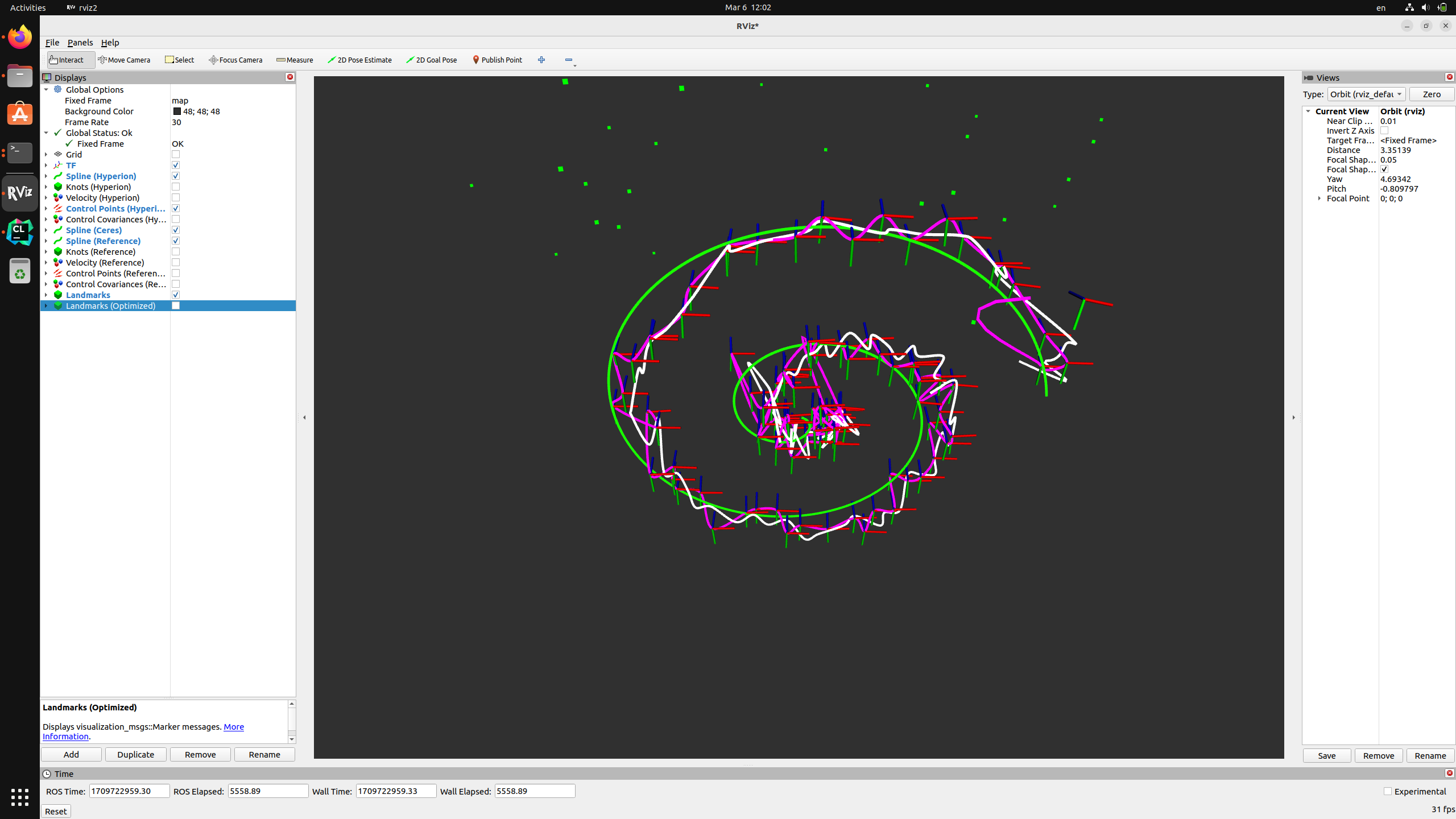}
    }
    \subfloat[Absolute Error]{
      \resizebox{!}{3.1cm}{\input{tables/visual_rmse}}
    }
    \caption{The motion estimates for \hyperion and a conventional \ac{NLLS} solver \cite{Agarwal:etal:Ceres} in magenta and white, respectively, converge to similar solutions close to ground truth (in green) in the localization setup, demonstrating robust convergence under poor pose ($\pm 0.20$ m/rad) and landmark initialization ($\pm 0.20$ m) with $\pm 1$ px measurement noise.}
    \label{fig:eccv2024:visual_setup}
\end{figure}

\begin{figure}
    \centering
    \subfloat[Convergence: Dropout]{
      \resizebox{!}{3.1cm}{\input{tables/visual_dropout}}
      \label{fig:eccv2024:visual_dropout}
    } 
    \subfloat[Convergence: Absolute]{
      \resizebox{!}{3.1cm}{\input{tables/absolute_energy}}
      \label{fig:eccv2024:absolute_convergence}
    } 
    \subfloat[Convergence: Localization]{
      \resizebox{!}{3.1cm}{\input{tables/visual_energy}}
      \label{fig:eccv2024:visual_convergence}
    }
    \caption{
        \subref{fig:eccv2024:visual_dropout} Graph energy vs. number of iterations conditioned on the dropout probability
in the localization setup. Convergence comparison of different splines and solver variants in the absolute setups \subref{fig:eccv2024:absolute_convergence} and the localization setup \subref{fig:eccv2024:visual_convergence}, respectively.
    }
\end{figure}
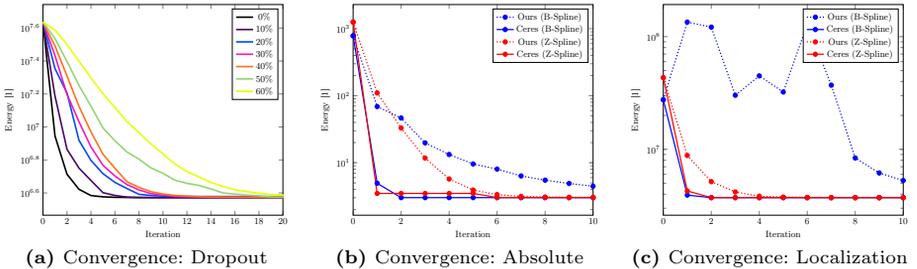

Overall, the results from the localization setup are consistent with our earlier observations, as detailed in \cref{fig:eccv2024:visual_setup,fig:eccv2024:visual_dropout,fig:eccv2024:visual_convergence}. However, \cref{fig:eccv2024:visual_dropout} indicates that this setup is more agnostic to lower dropout ratios compared to the absolute case. Furthermore, the disparity between different motion parameterizations becomes more pronounced over the previous setup as may be observed from \cref{fig:eccv2024:visual_convergence}. In general, visual setups also display higher volatility over absolute ones, with notable differences in estimates, especially towards the head and tails of the optimized splines. In a similar vein, we found that these numerical instabilities tend to have a destabilizing effect in loop graphs, posing considerable challenges to deploying the proposed system as a comprehensive \ac{SLAM} system at present.

\subsection{Performance Analysis}
As highlighted in \cref{sec:eccv2024:system_symbolic_factors}, our framework leverages SymForce \cite{Martiros:etal:RSS22} to automate the generation of analytic cost factors, aiming to outperform hand-crafted implementations while mitigating development effort and programmatic errors. Thus, we test our auto-generated B-Spline implementation against Sommer \etal's optimized, hand-crafted version \cite{Sommer:etal:CVPR2020}. The comparison, detailed in \cref{tab:eccv2024:timings}, assesses the timings of pose, velocity, and acceleration evaluations for B-Splines of order $k$, within a Lie group $\mathcal{L}$, potentially including the Jacobians $\partial / \partial \mathcal{B}$ with respect to the bases $\mathcal{B}_i$. \Cref{tab:eccv2024:timings} shows that our automated spline implementations significantly outperform the manual ones from \cite{Sommer:etal:CVPR2020}, with speedups ranging from 2.43x to 110.31x. The performance gains are particularly notable in derivative evaluations, where \cite{Sommer:etal:CVPR2020} employs automatic differentiation to compute the Jacobians. Considering the long-standing computational challenges of \ac{CTSLAM} methods, our automated approach not only reduces complexity but also narrows the performance divide with discrete-time \ac{SLAM}.

We further analyze the single-core performance of the proposed \ac{GBP} solver with Ceres \cite{Agarwal:etal:Ceres}, a conventional \ac{NLLS} solver, within the previously discussed scenarios. For the absolute setup (400 pose measurements), Ceres averages 0.70 ms per iteration, whereas our method averages 4.18 ms to update all graph vertices (\ie about 6x slower). In the localization scenario (10,000 landmark reprojections), Ceres requires 15 ms per iteration, compared to our solver's 112 ms per iteration (\ie 7.5x slower). This performance gap is, however, largely mitigated by considering that Ceres excludes the (implicit) estimation of covariances, significantly reducing computational load, and that it has been extensively refined by numerous contributors over the years. Moreover, the intrinsic distributedness and parallelizability of \ac{GBP} along with potential enhancements from adaptive vertex updates, leveraging covariance estimates to trigger selective updates, are promising avenues for further performance improvements. The above analysis suggests that, even in its present state, the proposed framework is capable of real-time execution for moderately sized problems.

%% file: tables/absolute_dropout.tex
\begin{tikzpicture}
    \pgfplotsset{
        scale only axis,
        xmin=0, xmax=20,
        colormap={bright}{rgb255=(0,0,0) rgb255=(78,3,100) rgb255=(2,74,255) rgb255=(255,21,181) rgb255=(255,113,26) rgb255=(147,213,114) rgb255=(230,255,0)
    rgb255=(255,255,255)}
    }
    \begin{semilogyaxis}[
      xlabel=Iteration,
      ylabel={Energy [1]},
      colormap name=bright, 
      cycle list={[of colormap]},
      every axis plot/.append style={mark=none,ultra thick}
    ]
    \legend{0\%, 10\%, 20\%, 30\%, 40\%, 50\%, 60\%}
    \addplot+ table [x=iter, y=drop_00, col sep=comma] {tables/absolute_dropout_energy.csv};
    \addplot+ table [x=iter, y=drop_10, col sep=comma] {tables/absolute_dropout_energy.csv};
    \addplot+ table [x=iter, y=drop_20, col sep=comma] {tables/absolute_dropout_energy.csv};
    \addplot+ table [x=iter, y=drop_30, col sep=comma] {tables/absolute_dropout_energy.csv};
    \addplot+ table [x=iter, y=drop_40, col sep=comma] {tables/absolute_dropout_energy.csv};
    \addplot+ table [x=iter, y=drop_50, col sep=comma] {tables/absolute_dropout_energy.csv};
    \addplot+ table [x=iter, y=drop_60, col sep=comma] {tables/absolute_dropout_energy.csv};
    \end{semilogyaxis}
\end{tikzpicture}

%% file: tables/charuco_energy.tex
\begin{tikzpicture}
    \pgfplotsset{
        scale only axis,
        xmin=0, xmax=10
    }
    \begin{semilogyaxis}[
      xlabel=Iteration,
      ylabel={Energy [1]},
    ]
    \legend{Ours, Ceres}
    \addplot[blue,mark=*,very thick,dotted,mark options={solid}] table [x=iter, y=cost, col sep=comma] {tables/hyperion_charuco_energy.csv};
    \addplot[very thick,mark=*] table [x=iter, y=cost] {tables/ceres_charuco_energy.txt};
    \end{semilogyaxis}
\end{tikzpicture}

%% file: tables/charuco_error.tex
\begin{tikzpicture}
    \pgfplotsset{
        scale only axis,
        xmin=0, xmax=60,
    }
    \begin{semilogyaxis}[
      axis y line*=left,
      xlabel=Time,
      ylabel={Relative Rotation Error [rad]},
      legend image post style={black},
      legend style = {text=black,font=\footnotesize},
    ]
    \addplot[very thick,smooth] table [x=t, y=dr, col sep=comma] {tables/hyperion_charuco_error.csv};
    \end{semilogyaxis}
    \begin{semilogyaxis}[
      blue,
      axis y line*=right,
      axis x line=none,
      ylabel={Relative Translation Error [m]}
    ]
    \addplot[very thick,smooth] table [x=t, y=dt, col sep=comma] {tables/hyperion_charuco_error.csv};
    \end{semilogyaxis}
\end{tikzpicture}

%% file: tables/visual_rmse.tex
\begin{tikzpicture}
    \pgfplotsset{
        scale only axis,
        xmin=0, xmax=9.9
    }
    \begin{axis}[
      axis y line*=left,
      xlabel=Time,
      ylabel={Absolute Rotation Error [rad]},
      legend pos=south west,
      legend image post style={black},
      legend style = {text=black,font=\footnotesize},
    ]
    \legend{Ours, Ceres}
    \addplot[very thick,smooth,dotted] table [x=t, y=dr, col sep=comma] {tables/hyperion_visual_rmse.csv};
    \addplot[very thick,smooth] table [x=t, y=dr, col sep=comma] {tables/ceres_visual_rmse.csv};
    \end{axis}
    \begin{axis}[
      blue,
      axis y line*=right,
      axis x line=none,
      ylabel={Absolute Translation error [m]}
    ]
    \addplot[very thick,smooth,dotted] table [x=t, y=dt, col sep=comma] {tables/hyperion_visual_rmse.csv};
    \addplot[very thick,smooth] table [x=t, y=dt, col sep=comma] {tables/ceres_visual_rmse.csv};
    \end{axis}
\end{tikzpicture}

%% file: tables/visual_dropout.tex
\begin{tikzpicture}
    \pgfplotsset{
        scale only axis,
        xmin=0, xmax=20,
        colormap={bright}{rgb255=(0,0,0) rgb255=(78,3,100) rgb255=(2,74,255) rgb255=(255,21,181) rgb255=(255,113,26) rgb255=(147,213,114) rgb255=(230,255,0)
    rgb255=(255,255,255)}
    }
    \begin{semilogyaxis}[
      xlabel=Iteration,
      ylabel={Energy [1]},
      colormap name=bright, 
      cycle list={[of colormap]},
      every axis plot/.append style={mark=none,ultra thick}
    ]
    \legend{0\%, 10\%, 20\%, 30\%, 40\%, 50\%, 60\%}
    \addplot+ table [x=iter, y=drop_00, col sep=comma] {tables/visual_dropout_energy.csv};
    \addplot+ table [x=iter, y=drop_10, col sep=comma] {tables/visual_dropout_energy.csv};
    \addplot+ table [x=iter, y=drop_20, col sep=comma] {tables/visual_dropout_energy.csv};
    \addplot+ table [x=iter, y=drop_30, col sep=comma] {tables/visual_dropout_energy.csv};
    \addplot+ table [x=iter, y=drop_40, col sep=comma] {tables/visual_dropout_energy.csv};
    \addplot+ table [x=iter, y=drop_50, col sep=comma] {tables/visual_dropout_energy.csv};
    \addplot+ table [x=iter, y=drop_60, col sep=comma] {tables/visual_dropout_energy.csv};
    \end{semilogyaxis}
\end{tikzpicture}

%% file: tables/absolute_energy.tex
\begin{tikzpicture}
    \pgfplotsset{
        scale only axis,
        xmin=0, xmax=10
    }
    \begin{semilogyaxis}[
      xlabel=Iteration,
      ylabel={Energy [1]},
    ]
    \legend{Ours (B-Spline), Ceres (B-Spline), Ours (Z-Spline), Ceres (Z-Spline)}
    \addplot[blue,mark=*,very thick,dotted,mark options={solid}] table [x=iter, y=cost, col sep=comma] {tables/hyperion_absolute_energy_bspline.csv};
    \addplot[blue,very thick,mark=*] table [x=iter, y=cost] {tables/ceres_absolute_energy_bspline.txt};
    \addplot[red,mark=*,very thick,dotted,mark options={solid}] table [x=iter, y=cost, col sep=comma] {tables/hyperion_absolute_energy_zspline.csv};
    \addplot[red,very thick,mark=*] table [x=iter, y=cost] {tables/ceres_absolute_energy_zspline.txt};
    \end{semilogyaxis}
\end{tikzpicture}

%% file: tables/visual_energy.tex
\begin{tikzpicture}
    \pgfplotsset{
        scale only axis,
        xmin=0, xmax=10
    }
    \begin{semilogyaxis}[
      xlabel=Iteration,
      ylabel={Energy [1]},
    ]
    \legend{Ours (B-Spline), Ceres (B-Spline), Ours (Z-Spline), Ceres (Z-Spline)}
    \addplot[blue,mark=*,very thick,dotted,mark options={solid}] table [x=iter, y=cost, col sep=comma] {tables/hyperion_visual_energy_bspline.csv};
    \addplot[blue,very thick,mark=*] table [x=iter, y=cost] {tables/ceres_visual_energy_bspline.txt};
    \addplot[red,mark=*,very thick,dotted,mark options={solid}] table [x=iter, y=cost, col sep=comma] {tables/hyperion_visual_energy_zspline.csv};
    \addplot[red,very thick,mark=*] table [x=iter, y=cost] {tables/ceres_visual_energy_zspline.txt};
    \end{semilogyaxis}
\end{tikzpicture}

%% file: sections/conclusions.tex
In this work, we present a fast, versatile \ac{GBP} framework that targets distributed, continuous-time \ac{SLAM} applications and leverages message-passing algorithms to achieve probabilistic inference. We demonstrate the efficacy and competitiveness of our method against a conventional \ac{NLLS} solver \cite{Agarwal:etal:Ceres} achieving similar convergence and performance properties in practical settings. We further provide a comprehensive library of high-performance implementations for continuous-time \ac{SLAM}, comprising motion parameterizations as well as common factors. Based on our experiments, the proposed framework shows great promise in paving the way towards resilient, distributed, continuous-time \ac{SLAM} solutions.

%% file: sections/acknowledgments.tex
This preprint has been accepted for publication in the European Conference on Computer Vision (ECCV) pending minor post-submission improvements and corrections. The Version of Record of this contribution will be published in the proceedings of the 18th European Conference on Computer Vision (ECCV 2024).\\


\noindent This work was partially funded by the European Research Council (ERC) Consolidator Grant project SkEyes (Grant agreement No. 101089328).